%% file: paper.tex
\documentclass{article} 
\usepackage{iclr2021_conference,times}

\usepackage{hyperref}
\usepackage{url}

\input{math_commands.tex}

\usepackage{url}
\usepackage{multicol}
\usepackage{subfig}
\usepackage{xspace}
\usepackage[textsize=small]{todonotes}
\usepackage{enumerate}

\newcommand{\srrns}{\textsc{sr-RNN}s\xspace}

\newcommand{\bfh}{\mathbf{h}}
\newcommand{\bfc}{\mathbf{c}}
\newcommand{\bfx}{\mathbf{x}}
\newcommand{\bfu}{\mathbf{u}}
\newcommand{\bfs}{\mathbf{s}}

\newcommand{\bfy}{\mathbf{y}}

\newcommand{\calQ}{\mathcal{Q}}
\newcommand{\calP}{\mathcal{P}}
\newcommand{\st}{ST-$\tau\,$}

\usepackage{amsmath, amssymb, amsthm}
\usepackage{algorithm}
\usepackage{algorithmic}
\usepackage{bbm}
\usepackage{multirow}
\usepackage{rotating}
\usepackage{placeins}
\usepackage{wrapfig}

\title{Uncertainty Estimation and Calibration with Finite-State Probabilistic RNNs}

\author{Cheng Wang\textsuperscript{*}, Carolin Lawrence\textsuperscript{*}, Mathias Niepert  \\
NEC Laboratories Europe\\
\texttt{\{cheng.wang,carolin.lawrence,mathias.niepert\}@neclab.eu} \\
}

\begingroup
\footnotetext{Equal contribution.}
\endgroup

\iclrfinalcopy 
\begin{document}

\maketitle

\begin{abstract}
Uncertainty quantification is crucial for building reliable and trustable machine learning systems. We propose to estimate uncertainty in recurrent neural networks (RNNs) via stochastic discrete state transitions over recurrent timesteps. The uncertainty of the model can be quantified by running a prediction several times, each time sampling from the recurrent state transition distribution, leading to potentially different results if the model is uncertain. Alongside uncertainty quantification, our proposed method offers several advantages in different settings. The proposed method can (1) learn deterministic and probabilistic automata from data, (2) learn well-calibrated models on real-world classification tasks, (3) improve the performance of out-of-distribution detection, and (4) control the exploration-exploitation trade-off in reinforcement learning.
\end{abstract}

\input{sec-introduction.tex}
\input{sec-srrnn.tex}
\input{sec-related.tex}
\input{sec-experiments.tex}

\input{sec-conclusion.tex}

\bibliography{iclr2021_conference}
\bibliographystyle{iclr2021_conference}

\clearpage

\appendix
\input{sec-appendix.tex}

\end{document}

%% file: math_commands.tex
\usepackage{amsmath,amsfonts,bm}

\def\eqref#1{equation~\ref{#1}}

\def\1{\bm{1}}

\DeclareMathAlphabet{\mathsfit}{\encodingdefault}{\sfdefault}{m}{sl}
\SetMathAlphabet{\mathsfit}{bold}{\encodingdefault}{\sfdefault}{bx}{n}

\DeclareMathOperator*{\argmax}{arg\,max}

%% file: sec-introduction.tex
\section{Introduction}

\begin{figure}
	\vspace{-5mm}
	\centering
	\subfloat[Heart Rate Classification]{{\includegraphics[width=0.49\textwidth]{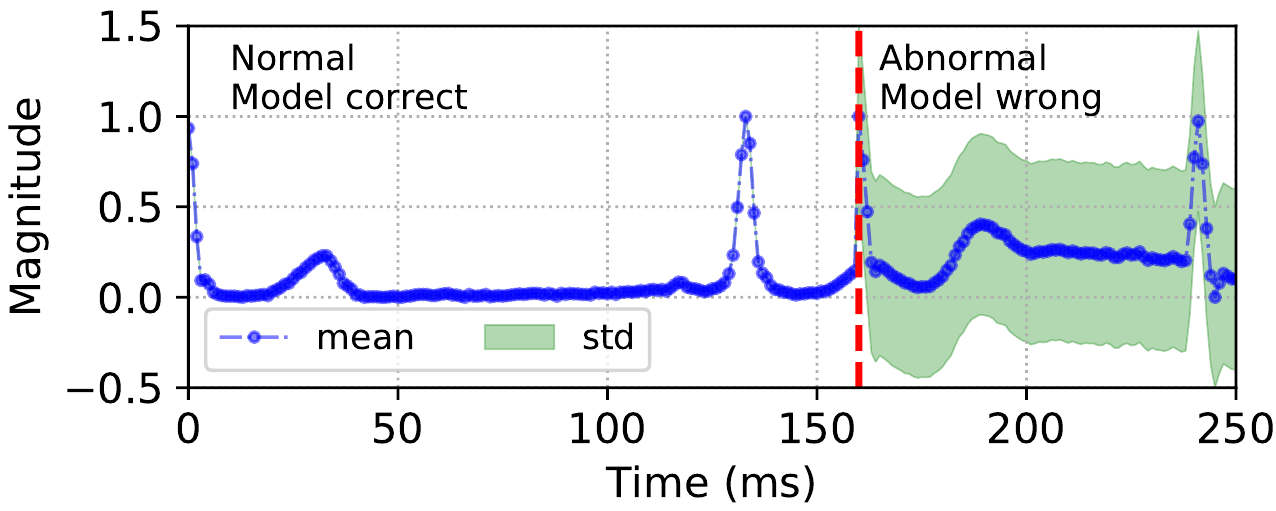} }} 
	\subfloat[Sentiment Classification]{{\includegraphics[width=0.497\textwidth]{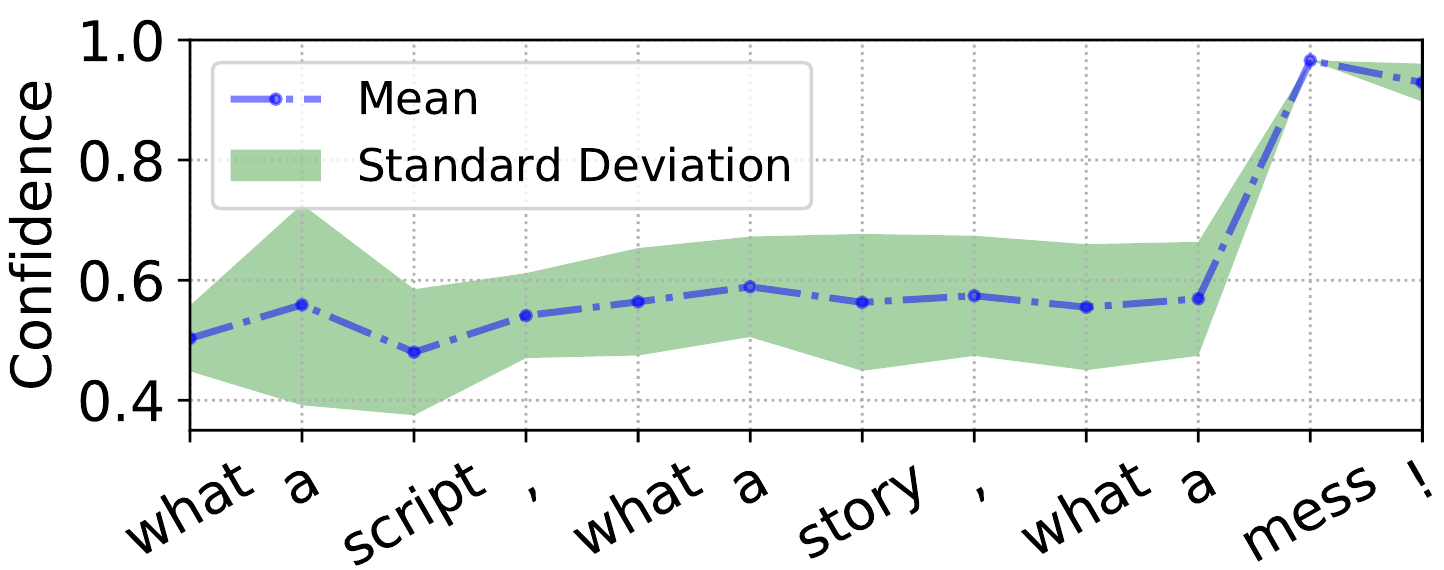} }}
	\caption{\label{fig:example_uncertain} (a) Prediction uncertainty of \st, our proposed method, for an ECG time-series based on 10 runs. To the left of the red line \st classifies a heart beat as normal. To the right of the red line, \st makes wrong predictions. Due to its drop in certainty, however, it can alert medical personnel. (b) Given a sentence with negative sentiment, \st reads the sentence word by word. The y-axis presents the model's confidence of the sentence having a negative sentiment. After the first few words, the model leans towards a negative sentiment, but is uncertain about its prediction. After the word ``\textit{mess},'' its uncertainty drops and it predicts the sentiment as negative.}
\end{figure}

Machine learning models are well-calibrated if the probability associated with the predicted class reflects its correctness likelihood relative to the ground truth. The output probabilities of modern neural networks are often poorly calibrated~\citep{guo2017calibration}. For instance, typical neural networks with a softmax activation tend to assign high probabilities to out-of-distribution samples~\citep{GalGhahramani:16b}.  Providing uncertainty estimates is important for model interpretability as it allows users to assess the extent to which they can trust a given prediction~\citep{jiang2018trust}. Moreover, well-calibrated output probabilities are crucial in several use cases. For instance, when monitoring medical time-series data (see Figure~\ref{fig:example_uncertain}(a)), hospital staff should also be alerted when there is a low-confidence prediction concerning a patient's health status. 

Bayesian neural networks (BNNs), which place a prior distribution on the model's parameters, are a popular approach to modeling uncertainty. BNNs often require more parameters, approximate inference, and depend crucially on the choice of prior~\citep{Gal:16,LakshminarayananETAL:17}. Applying dropout both during training and inference can be interpreted as a BNN and provides a more efficient method for uncertainty quantification \citep{GalGhahramani:16b}. The dropout probability, however, needs to be tuned and, therefore, leads to a trade-off between predictive error and calibration error.

Sidestepping the challenges of Bayesian NNs, we propose an orthogonal approach to quantify the uncertainty in recurrent neural networks (RNNs). At each time step, based on the current hidden (and cell) state, the model computes a probability distribution over a finite set of states. The next state of the RNN is then drawn from this distribution. We use the Gumbel softmax trick~\citep{Gumbel:54,KendallGal:17,JangETAL:17} to perform Monte-Carlo gradient estimation. Inspired by the effectiveness of temperature scaling~\citep{guo2017calibration} which  is usually applied to trained models, we learn the temperature $\tau$ of the Gumbel softmax distribution during training to control the concentration of the state transition distribution. Learning $\tau$ as a parameter can be seen as entropy regularization \citep{ szegedy2016rethinking, pereyra2017regularizing,JangETAL:17}. The resulting model, which we name \st, defines for every input sequence a probability distribution over state-transition paths similar to a probabilistic state machine. To estimate the model's uncertainty for a prediction, \st is run multiple times to compute mean and variance of the prediction probabilities. 

We explore the behavior of \st in a variety of tasks and settings. First, we show that \st can learn deterministic and probabilistic automata from data. Second, we demonstrate on real-world classification tasks that \st learns well calibrated models. Third, we show that \st is competitive in out-of-distribution detection tasks. Fourth, in a reinforcement learning task, we find that \st is able to trade off exploration and exploitation behavior better than existing methods. Especially the out-of-distribution detection and reinforcement learning tasks are not amenable to post-hoc calibration approaches \citep{guo2017calibration} and, therefore, require a method such as ours that is able to calibrate the probabilities during training. 

%% file: sec-srrnn.tex
\section{Uncertainty in Recurrent Neural Networks}
\subsection{Background}
An RNN is a function $f$ defined through a neural network with parameters $\mathbf{w}$ that is applied over time steps: at time step $t$, it reuses the hidden state $\mathbf{h}_{t-1}$ of the previous time step and the current  input $\mathbf{x}_{t}$ to compute a new state $\mathbf{h}_{t}$, $f : (\mathbf{h}_{t-1}, \mathbf{x}_{t}) \rightarrow \mathbf{h}_{t}$. Some RNN variants such as LSTMs have memory cells $\mathbf{c}_t$ and apply the function $f : (\mathbf{h}_{t-1}, \mathbf{c}_{t-1}, \mathbf{x}_{t}) \rightarrow \mathbf{h}_{t}$ at each step. A vanilla RNN maps two identical input sequences to the same state and it is therefore not possible to measure uncertainty of a prediction by running inference multiple times. Furthermore, it is known that passing $\mathbf{h}_t$ through a softmax transformation leads to overconfident predictions on out-of-distribution samples and poorly calibrated probabilities \citep{guo2017calibration}. In a Bayesian RNN the weight matrices $\mathbf{w}$ are drawn from a distribution and, therefore, the output is an average of an infinite number of models. Unlike vanilla RNNs, Bayesian RNNs are stochastic and it is possible to compute average and variance for a prediction. Using a prior to integrate out the parameters during training also leads to a regularization effect. However, there are two major and often debated challenges of BNNs: the right choice of prior and the efficient approximation of the posterior.

\begin{figure}
    \vspace{-5mm}
	\centering
	\includegraphics[width=0.85\textwidth]{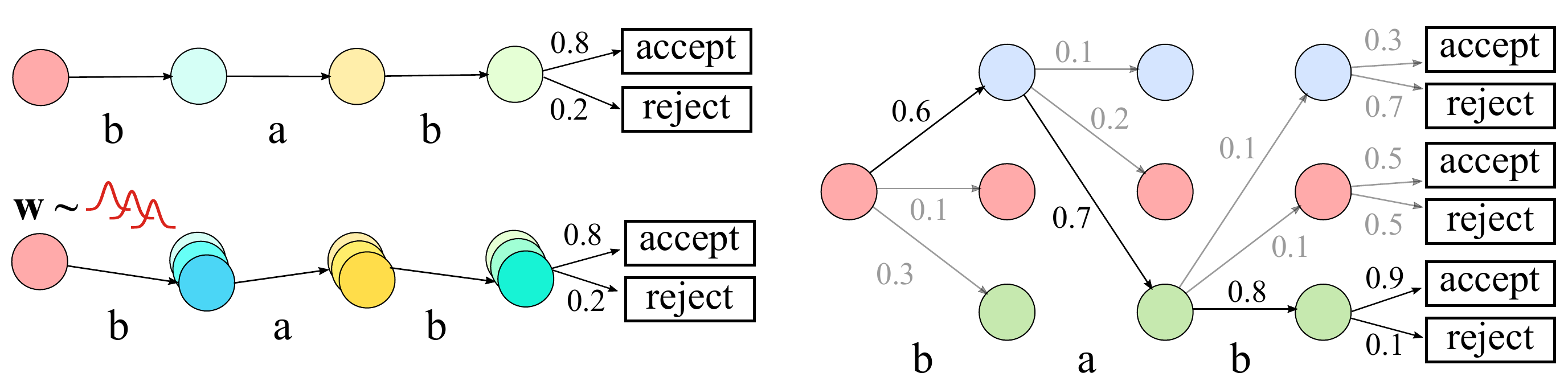}
	\caption{\label{fig:example-paths} Illustration of several ways to model uncertainty in recurrent models. Left, top: In standard recurrent models the sequence of hidden states is identical for any two runs on the same input. Uncertainty is typically modeled with a softmax distribution over the classes (here: accept and reject). Left, bottom: Bayesian NNs make the assumption that the weights are drawn from a distribution. Uncertainty is estimated through model averaging. Right: The proposed class of RNNs assumes that there is a \emph{finite} number of states between which the RNN transitions. Uncertainty for an input sequence is modeled through a probability distribution over possible state-transition paths.}
\end{figure}

With this paper, we side-step these challenges and model the uncertainty of an RNN through probabilistic state transitions between a finite number of $k$ learnable states $\mathbf{s_1}, ..., \mathbf{s}_k$. Given a state $\mathbf{h}_{t}$, we compute a stochastic probability distribution over the learnable states. Hence, for the same state and input, the RNN might move to different states in different runs. Instead of integrating over possible weights, as in the case of BNNs, we sum over all possible state sequences and weigh the classification probabilities by the probabilities of these sequences. Figure~\ref{fig:example-paths} illustrates the proposed approach and contrasts it with vanilla and Bayesian RNNs.
The proposed method combines two building blocks. The first is state-regularization \citep{WangNiepert:19} as a way to compute a probability distribution over a finite set of states in an RNN. State-regularization, however, is  deterministic and therefore we utilize the second building block, the Gumbel softmax trick \citep{Gumbel:54,MaddisonETAL:17,JangETAL:17} to sample from a categorical distribution. Combining the two blocks allows us to create a stochastic state RNN which can model uncertainty. Before we formulate our method, we first introduce the necessary two building blocks.

\paragraph{Deterministic State-Regularized RNNs.} 
State regularization \citep{WangNiepert:19} extends RNNs by dividing the computation of the hidden state $\bfh_{t}$ into two components. The first component is an intermediate vector $\bfu_{t} \in \mathbb{R}^{d}$ computed in the same manner as the standard recurrent component, $\bfu_{t} = f\left(\bfh_{t-1}, \bfc_{t-1},\bfx_{t}\right)$. The second component models probabilistic state transitions between a finite set of $k$ learnable states $\bfs_1$, \dots, $\bfs_k$, where $\bfs_i$ $\in$ $\mathbb{R}^{d}, i \in [1,k]$ and which can also be written as a matrix $\mathbf{S}_t \in \mathbb{R}^{d \times k}$. 
$\mathbf{S}_t$ is randomly initialized and learnt during backpropagation like any other network weight. 
At time step $t$, given an $\bfu_{t}$, the transition over next possible states is computed by: $\bm{\theta}_t =\varphi(\mathbf{S}_t, \bfu_t)$, where $\bm{\theta}_t=\{\theta_{t,1},..., \theta_{t,k}\}$ and $\varphi$ is some pre-defined function. In \citet{WangNiepert:19}, $\varphi$ was a matrix-vector product followed by a \textsc{Softmax} function that ensures $\sum_{i=1}^{k} \theta_{t, i}=1$. The hidden state $\bfh_t$ is then computed by
\begin{equation}\label{eq:state}
\bfh_t = g(\bm{\theta}_t) \cdot \mathbf{S}_t^\top, ~~~\bfh_t  \in \mathbb{R}^d,
\end{equation}
where $g(\cdot)$ is another function, e.g. to compute the average. Because Equation (\ref{eq:state}) is deterministic, it cannot capture and estimate epistemic uncertainty.

\paragraph{Monte-Carlo Estimator with Gumbel Trick.} The Gumbel softmax trick is an instance of a path-wise Monte-Carlo gradient estimator~\citep{Gumbel:54,MaddisonETAL:17,JangETAL:17}. With the Gumbel trick, it is possible to draw samples $z$ from a categorical distribution given by paramaters $\bm{\theta}$, that is, 
$\bm{z}= \mbox{\textsc{one\_hot}} \big( \argmax_i [\gamma_i + \log\theta_i] \big), i \in [1\dots k]$,
where $k$ is the number of categories and $\gamma_i$ are i.i.d. samples from the \textsc{Gumbel}$(0, 1)$, that is, $\gamma=-\log(-\log(u)), u \sim \textsc{uniform}(0, 1)$. Because the $\argmax$ operator breaks end-to-end differentiability, the categorical distribution $\bm{z}$ can be approximated using the differentiable softmax function \citep{JangETAL:17,MaddisonETAL:17}. This enables us to draw a $k$-dimensional sample vector $\bm{\alpha}\in\Delta^{k-1}$, where $\Delta^{k-1}$ is the $(k-1)$-dimensional probability simplex.

\subsection{Stochastic Finite-State RNNs (\st)}
Our goal is to make state transitions stochastic and uitilize them to measure uncertainty: given an input sequence, the uncertainty is modeled via the probability distribution over possible state-transition paths (see right half of Figure \ref{fig:example-paths}). We can achieve this by setting $\varphi$ to be a matrix-vector product and using $\bm{\theta}_t$ to sample from a Gumbel softmax distribution with temperature parameter $\tau$. 
Applying Monte Carlo estimation, at each time step $t$, we sample a distribution over state transition probabilities $\bm{\alpha}_t$ from the Gumbel softmax distribution with current parameter $\tau$, where each state transition has the probability
\begin{equation}
\alpha_{t, i} = \frac{\exp ((\log(\theta_{t, i}) +\gamma_i)/\tau)}{\sum_{j=1}^{k} \exp ((\log(\theta_{t, j})+\gamma_j)/\tau)}, i \in [1\dots k].
\end{equation}

The resulting $\bm{\alpha}_t= \{\alpha_{t,1}, ..., \alpha_{t,k}\}$ can be seen as a probability distribution that judges how important each learnable state  $\bfs_{t}$ is. 
The new hidden state $\bfh_t$ can now be formed either as an average, $\bfh_t = \sum_{i=1}^{k} \alpha_{t,i} \bfs_i$ (the ``soft" Gumbel estimator), or as a one-hot vector, $\bfh_t= \mbox{\textsc{one\_hot}} \big(\argmax_i [\log(\theta_{t, i})+\gamma_i] \big)$. 
For the latter, gradients can be estimated using the straight-through estimator. Empirically, we found the average to work better. By sampling from the Gumbel softmax distribution at each time step, the model is stochastic and it is possible to measure variance across predictions and, therefore, to estimate epistemic uncertainty. For more theoretical details we refer the reader to Appendix~\ref{app:uncertainty}.

The parameter $\tau$ of the Gumbel softmax distribution is learned during training~\citep{JangETAL:17}. This allows us to directly adapt probabilistic RNNs to the inherent uncertainty of the data. Intuitively, the parameter $\tau$ influences the concentration of the categorical distribution, that is, the larger $\tau$ the more uniform the distribution. Since we influence the state transition uncertainty with the learned temperature $\tau$, we refer to our model as \st. We provide an ablation experiment of learning $\tau$ versus keeping it fixed in Appendix \ref{app:ablation}. 

Figure \ref{fig:st_overview} illustrates the proposed model. Given the previous hidden state of an RNN, first an intermediate hidden state $\bfu_{t}$ is computed using a standard RNN cell. Next, the intermediate representation $\bfu_{t}$ is multiplied with $k$ learnable states arranged as a matrix $\mathbf{S}_t$, resulting in $\bm{\theta}_t$. Based on $\bm{\theta}_t$, samples are drawn from a Gumbel softmax distribution with learnable temperature parameter $\tau$. The sampled probability distribution represents the certainty the model has in moving to the other states. Running the model on the same input several times (drawing Monte-Carlo samples) allows us to estimate the uncertainty of the \st model. 

\begin{figure}[t!]
	\vspace{-5mm}
	\centering
	\includegraphics[width=0.9\textwidth]{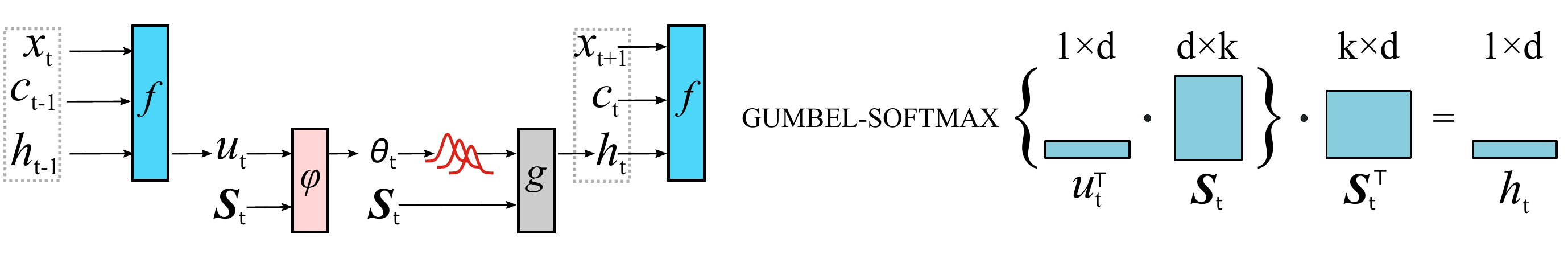}
	\caption{\label{fig:st_overview} One step of \st (for batch size $b=1$). First, previous hidden state $\mathbf{h}_{t-1}$, previous cell state $\mathbf{c}_{t-1}$ (if given) and input $\mathbf{x}_t$, are passed into an RNN cell (e.g. LSTM). The RNN cell returns an updated hidden state $\mathbf{u}_t$ and cell state $\mathbf{c}_t$. Second, $\mathbf{u}_t$ is further processed by using the learnable states matrix $\mathbf{S}_t$ in the state transition step (Right) and returns a new hidden state $\bm{h}_t$.}
\end{figure}

\subsection{Aleatoric and Epistemic Uncertainty}\label{app:uncertainty}
Let $\mathcal{Y}$ be a set of class labels and $\mathcal{D}$ be a set of training samples. For a classification problem and a given \st model with states $\{\mathbf{s}_1,... , \mathbf{s}_k\}$, we can write for every $y \in \mathcal{Y}$ 
\begin{equation}
\label{eqn-path-distribution}
p(y \mid \mathbf{x}=\langle\mathbf{x}_1, ..., \mathbf{x}_n\rangle) = \sum_{\langle\mathbf{h}_{1},...,\mathbf{h}_{n}\rangle \in \Psi}  p(\langle\mathbf{h}_{1},...,\mathbf{h}_{n}\rangle \mid  \mathbf{x})\ q\left(y \mid \mathbf{h}_n \right)
\end{equation}
where $\mathbf{h}_i \in \{\mathbf{s}_1,... , \mathbf{s}_k\}$ and the sum is over all possible paths (state sequences) $\Psi$ of length $n$. Moreover, $p(\langle\mathbf{h}_{1},...,\mathbf{h}_{n}\rangle \mid \mathbf{x})$ is the probability of path $\psi = \langle\mathbf{h}_{1},...,\mathbf{h}_{n}\rangle \in \Psi$ given input sequence $\mathbf{x}$ and $q\left(y \mid \mathbf{h}_n \right)$ is the probability of class $y$ given that we are in state $\mathbf{h}_n$. 
Instead of integrating over possible weights, as in the case of BNNs, with \st we integrate (sum) over all possible paths and weigh the class probabilities by the path probabilities.
The above model implicitly defines a probabilistic ensemble of several deterministic models, each represented by a particular path. As mentioned in a recent paper about aleatoric and epistemic uncertainty in ML~\citep{hullermeier2019aleatoric}: ``the variance of the predictions produced by an ensemble is a good indicator of the (epistemic) uncertainty in a prediction." 

Let us now make this intuition more concrete using recently proposed measures of aleatoric and epistemic uncertainty ~\citep{depeweg2018decomposition}; further discussed in \citep{hullermeier2019aleatoric}. 
In Equation (19) of \citep{hullermeier2019aleatoric} the \emph{total} uncertainty is defined as the entropy of the predictive posterior distribution $$H \left[ p(y \mid \mathbf{x}) \right] = - \sum_{y \in \mathcal{Y}} p(y \mid \mathbf{x}) \log_2 p(y | \mathbf{x}).$$ The above term includes both aleatoric \emph{and} epistemic uncertainty~\citep{depeweg2018decomposition,hullermeier2019aleatoric}. Now, in the context of Bayesian NNs, where we have a distribution over the weights of a neural network, the expectation of the entropies wrt said distribution is the aleatoric uncertainty (Equation 20 in \cite{hullermeier2019aleatoric}):
$$\mathbf{E}_{p(\mathbf{w} \mid \mathcal{D})} H\left[ p(y \mid \mathbf{w}, \mathbf{x})\right] = -\int p(\mathbf{w} \mid \mathcal{D})\left(\sum_{y \in \mathcal{Y}} p(y \mid \mathbf{w},\mathbf{x})\log_2 p(y \mid \mathbf{w},\mathbf{x}) \right) d\mathbf{w}.$$
Fixing the parameter weights to particular values eliminates the epistemic uncertainty. Finally, the epistemic uncertainty is obtained as
the difference of the total and aleatoric uncertainty (Equation 21 in \cite{hullermeier2019aleatoric}):
$$u_e(\mathbf{x}) := H \left[ p(y \mid \mathbf{x}) \right] - \mathbf{E}_{p(\mathbf{w} \mid \mathcal{D})} H\left[ p(y \mid \mathbf{w}, \mathbf{x})\right].$$

\begin{figure*}[t!]
	\centering
	\includegraphics[width=0.7\textwidth]{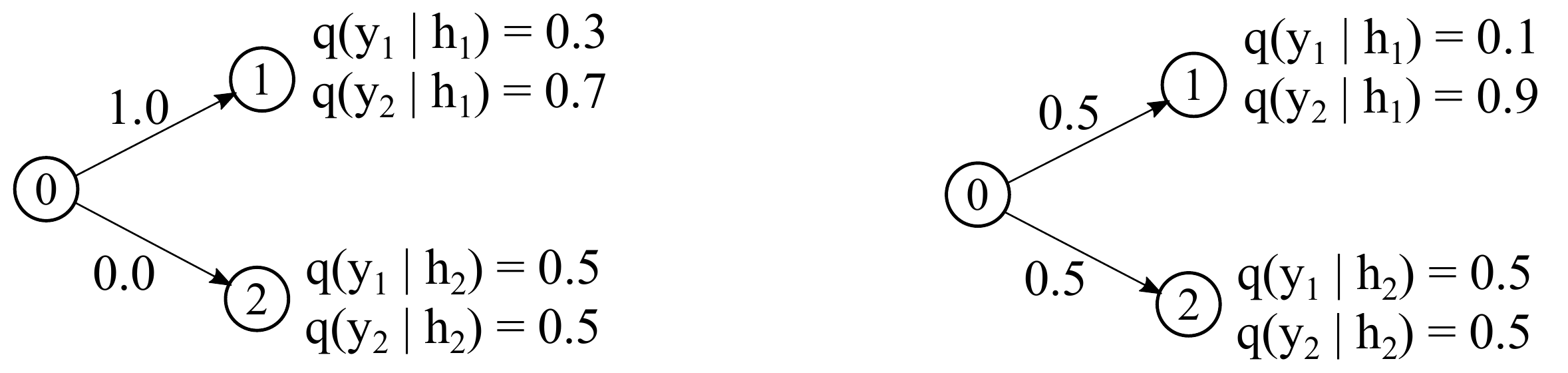}
\caption{\label{epistemic-exp}Two different probabilistic finite-state RNNs.}
\end{figure*}

Now, let us return to finite-state probabilistic RNNs. Here, the aleatoric uncertainty is the expectation of the entropies with respect to the distribution over the possible paths $\Psi$:
$$\mathbf{E}_{p(\psi \mid \mathbf{x})} H\left[ p(y \mid \psi, \mathbf{x})\right] = -\sum_{\psi \in \Psi} p(\psi \mid \mathbf{x})\left(\sum_{y \in \mathcal{Y}} p(y \mid \psi,\mathbf{x})\log_2 p(y \mid \psi,\mathbf{x}) \right),$$
where $p(y \mid \psi,\mathbf{x})$ is the probability of class $y$ conditioned on $\mathbf{x}$ and a particular path $\psi$. 
The epistemic uncertainty for probabilistic finite-state RNNs can then be computed by
$$u_e(\mathbf{x}) := H \left[ p(y \mid \mathbf{x}) \right] - \mathbf{E}_{p(\psi \mid \mathbf{x})} H\left[ p(y \mid \psi, \mathbf{x})\right].$$
Probabilistic finite-state RNNs capture epistemic uncertainty when the equation above is non-zero. As an example let us take a look at the two \st models given in Figure~\ref{epistemic-exp}. 
Here, we have for input $\mathbf{x}$ two class labels $y_1$ and $y_2$, three states ($\mathbf{s}_i, i\in \{0, 1, 2\}$), and two paths. In both cases, we have that $H \left[ p(y \mid \mathbf{x}) \right] = - \left( 0.3 \log_2 0.3 + 0.7 \log_2 0.7\right) \approx 0.8813.$
Looking at the term for the aleatoric uncertainty, for the \st depicted on the left side we have 
$\mathbf{E}_{p(\psi \mid \mathbf{x})} H\left[ p(y \mid \psi, \mathbf{x})\right] = - \left( 0.3 \log_2 0.3 + 0.7 \log_2 0.7\right) \approx 0.8813.$
In contrast, for the \st depicted on the right side we have $\mathbf{E}_{p(\psi \mid \mathbf{x})} H\left[ p(y \mid \psi, \mathbf{x})\right] \approx 0.7345.$ 
Consequently, the left side \st has an epistemic uncertainty of $u_e(\mathbf{x})=0$ but the right side \st exhibits an epistemic uncertainty of $u_e(\mathbf{x})=0.1468$.

This example illustrates three main observations. First, we can represent epistemic uncertainty through distributions over possible paths. Second, the more spiky the transition distributions, the more deterministic the behavior of the \st model, and the more confident it becomes with its prediction by shrinking the reducible source of uncertainty (epistemic uncertainty). Third, both models are equally calibrated as their predictive probabilities are, in expectation, identical for the same inputs. Hence, \st is not merely calibrating predictive probabilities but also captures epistemic uncertainty. Finally, we want to stress the connection between the parameter $\tau$ (the temperature) and the degree of (epistemic) uncertainty of the model. For small $\tau$  the \st model behavior is more deterministic and, therefore, has a lower degree of epistemic uncertainty. For instance, the \st on the left in Figure~\ref{epistemic-exp} has $0$ epistemic uncertainty because all transition probabilities are deterministic. Empirically, we find that the temperature and, therefore, the epistemic uncertainty often reduces during training, leaving the irreducible uncertainty (aleatoric) to be the main source of uncertainty. 

%% file: sec-related.tex
\section{Related Work}

\textbf{Uncertainty}.
Uncertainty quantification for safety-critical applications \citep{KrzywinskiAltman:13} has  been explored for deep neural nets in the context of Bayesian learning \citep{BlundellETAL:15,GalGhahramani:16b,KendallGal:17,KwonETAL:18}.
Bayes by Backprop (BBB) \citep{BlundellETAL:15} is a variational inference scheme for learning a distribution over weights $\mathbf{w}$ in neural networks and assumes that the weights are distributed normally, that is, $w_i \sim \mathcal{N}(\mu, \sigma^2)$. The principles of bayesian neural networks (BNNs) have been applied to RNNs and shown to result in superior performance compared to vanilla RNNs in natural language processing (NLP) tasks~\citep{FortunatoETAL:17}. However, BNNs come with a high computational cost because we need to learn $\mu$ and $\sigma$ for each weight in the network, effectively doubling the number of parameters. Furthermore, the prior might not be optimal and approximate inference could lead inaccurate estimates \citep{KuleshovETAL:18}. 
Dropout \citep{HintonETAL:12,SrivastavaETAL:14} can be seen as a variational approximation of a Gaussian Process~\citep{GalGhahramani:16b,GalGhahramani:16}. By leaving dropout activated  at prediction time, it can be used to measure uncertainty. However, the dropout probability needs to be tuned which leads to a trade-off between predictive error and calibration error (see Figure \ref{fig:vd} in the Appendix for an empirical example). 
Deep ensembles~\citep{LakshminarayananETAL:17} offer a non-Bayesian approach to measure uncertainty by training multiple separate networks and ensembling them. Similar to BNNs, however, deep ensembles require more resources as several different RNNs need to be trained. We show that \st is competitive to deep ensembles without the resource overhead. Recent work~\citep{hwang2020sampling} describes a sample-free uncertainty estimation for Gated Recurrent Units (SP-GRU)~\citep{chung2014empirical}, which estimates uncertainty by performing forward propagation in a series of deterministic linear and nonlinear transformations with exponential family distributions. \st estimates uncertainties through the stochastic transitions between two consecutive recurrent states. 

\textbf{Calibration}. Platt scaling \citep{platt1999probabilistic} is a calibration method for binary classification settings and has been extend to multi-class problems \citep{zadrozny2002transforming} and the structured prediction settings~\citep{kuleshov2015calibrated}.  \citep{guo2017calibration} extended the method to calibrate modern deep neural networks, particularly networks with a large number of layers. In their setup, a temperature parameter for the final softmax layer is adjusted only after training. In contrast, our method learns the  temperature and, therefore, the two processes are not decoupled. In some tasks such as time-series prediction or RL it is crucial to calibrate during training and not post-hoc.

\textbf{Deterministic \& Probabilistic Automata Extraction}.
Deterministic Finite Automata (DFA) have been used to make the behavior of RNNs more transparent.  
DFAs can be extracted from RNNs after an RNN is trained by applying clustering algorithms like $k$-means to the extracted hidden states~\citep{Wang:2007:nc} or by applying the exact learning algorithm L$^*$ \citep{pmlr-v80-weiss18a}. Post-hoc extraction, however, might not recover faithful DFAs. Instead, \citep{WangNiepert:19} proposed state-regularized RNNs where the finite set of states is learned alongside the RNN by using probabilistic state transitions.
Building on this, we use the Gumbel softmax trick to model stochastic state transitions, allowing us to learn probabilistic automata (PAs)~\citep{rabin1963probabilistic} from data.

\textbf{Hidden Markov Models (HMMs) \& State-Space Models (SSMs) \& RNNs}.
HMMs are transducer-style probabilistic automata, simpler and more transparent models than RNNs. \citep{bridle1990alpha} explored how a HMM can be interpreted as an RNNs by using full likelihood scoring for each word model. \citep{krakovna2016increasing} studied various combinations of HMMs and RNNs to increase the interpretability of RNNs. There have also been ideas on incorporating RNNs to HMMs to capture complex dynamics~\citep{dai2016recurrent,doerr2018probabilistic}. Another relative group of work is SSMs, e.g., rSLDS~\citep{linderman2017bayesian}, Kalman VAE~\citep{fraccaro2017disentangled}, PlaNet~\citep{hafner2019learning} and RKN~\citep{BeckerPandyaGebhardt2019_1000118248}. They can be extended and viewed as another way to inject stochasticity to RNN-based architectures. 
In contrast, \st models stochastic finite-state transition mechanisms end-to-end in conjunction with modern gradient estimators to directly quantify and calibrate uncertainty and the underlying probabilistic system. This enables \st to approximate and extract the probabilistic dynamics in RNNs. 

%% file: sec-experiments.tex
\section{Experiments}
The experiments are grouped into five categories. First, we show that it is possible to use \st to learn deterministic and probabilistic automata from language data (Sec. \ref{sec:pfsm}). This demonstrates that \st can capture and recover the stochastic behavior of both deterministic and stochastic languages. 
Second, we demonstrate on classification tasks (Sec. \ref{sec:class}) that \st performs better than or similar to existing baselines both in terms of predictive quality and model calibration. Third, we compare \st with existing baselines using out-of-distribution detection tasks (Sec. \ref{subsection-ood-experiments}). 
Fourth, we conduct reinforcement learning experiments where we show that the learned parameter $\tau$ can calibrate the exploration-exploitation trade-off during learning (Appendix \ref{sec:rl}), leading to a lower sample complexity. Fifth, we report results on a regression task (Appendix \ref{sec:reg}).

\paragraph{Models.} We compare the proposed \st method to four existing models. 
First, a vanilla LSTM (\textbf{LSTM}). Second, a Bayesian RNN (\textbf{BBB})~\citep{BlundellETAL:15, FortunatoETAL:17}, where each network weight $w_i$ is sampled from a Gaussian distribution $\mathcal{N} (\mu,\sigma^2)$, $w_i= \mu+\sigma \cdot \epsilon $, with $\sigma=\log(1+\exp(\rho))$ and $\epsilon \sim \mathcal{N}(0, 1)$. To estimate model uncertainty, we keep the same sampling strategy as employed during training (rather than using $w_i=\mu$). 
Third, a RNN that employs Variational Dropout (\textbf{VD}) \citep{GalGhahramani:16}. In variational dropout, the same dropout mask is applied at all time steps for one input. To use this method to measure uncertainty, we keep the same dropout probability at prediction time ~\citep{GalGhahramani:16}. Fourth, a deep ensemble of a LSTM base model~\citep{LakshminarayananETAL:17}. For \st we compute the new hidden state using the soft version of the Gumbel softmax estimator~\citep{JangETAL:17}.
All models contain only a single LSTM layer, are implemented in Tensorflow \citep{45166}, and use the \textsc{Adam} \citep{adam} optimizer with initial learning rate $0.001$.

\subsection{Deterministic \& Probabilistic Automata Extraction}\label{sec:pfsm}
We aim to investigate the extent to which \st can learn deterministic and probabilistic automata from sequences generated by regular and stochastic languages. Since the underlying languages are known and the data is generated using these languages, we can exactly assess whether \st can recover these languages. 
We refer the reader to the appendix~\ref{app:dfa_pa} for a definition of deterministic finite automata (DFA) and probabilistic automata (PA).  The set of languages recognized by DFAs and PAs are referred to as, respectively, \textit{regular} and \textit{stochastic} languages. For the extraction experiments we use the GRU cell as it does not have a cell state. This allows us to read the Markovian transition probabilities for each state-input symbol pair directly from the trained \st. 

\begin{figure}
	\vspace{-10mm}
	\centering
	\subfloat[\scriptsize True PA-1]{{\includegraphics[width=0.22\textwidth]{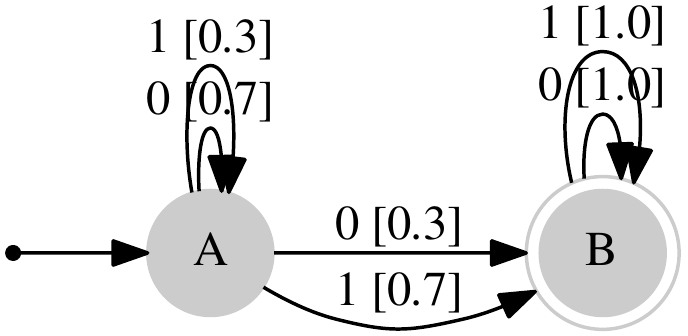} } }
    \hspace{2mm}
	\subfloat[\scriptsize Extracted PA-1]{{\includegraphics[width=0.22\textwidth]{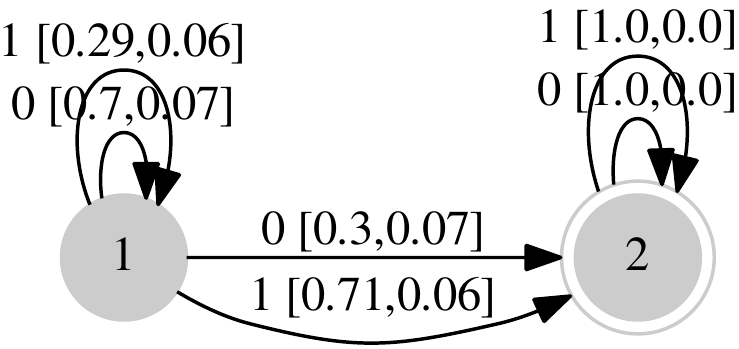} }}
    \hspace{10mm}
    \subfloat[\scriptsize True PA-2]{{\includegraphics[width=0.2\textwidth]{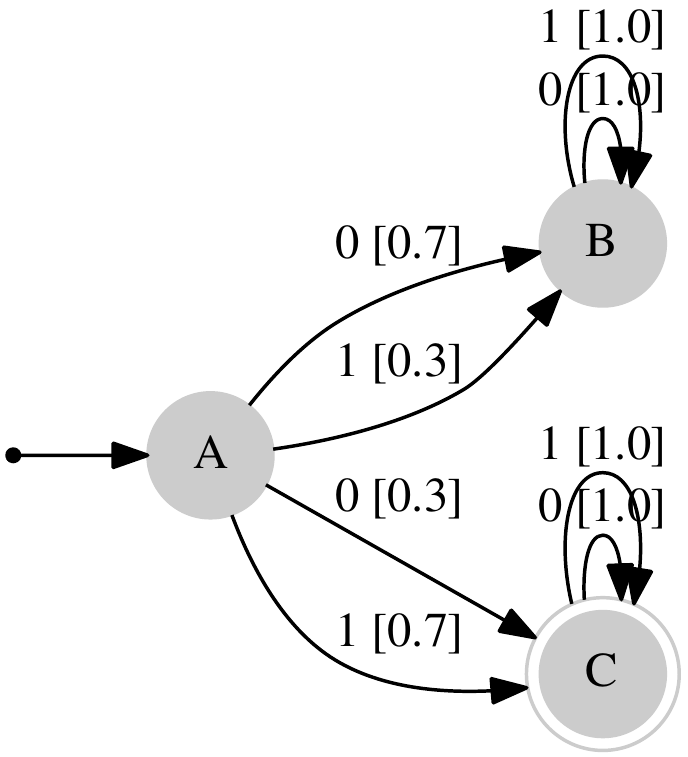} }}
    \hspace{2mm}
	\subfloat[\scriptsize Extracted PA-2]{{\includegraphics[width=0.18\textwidth]{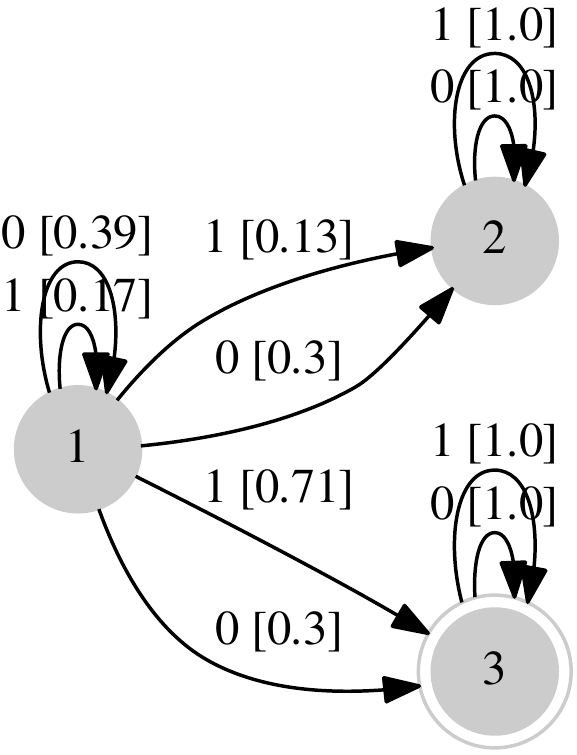} }}
	\caption{\label{fig:pfsm_ndfa} (a)+(c): ground truth probabilistic automata (PAs) used for generating training samples. (b)+(d) the PAs extracted from \st.The single and double circles represent \textit{reject} and \textit{accept} states respectively. The PA (d) resulted from a minimization of the actual extracted PA.}
\end{figure} 

\textbf{Regular Languages}.
We conduct experiments on the regular language defined by Tomita grammar $3$. This language consists of any string without an odd number of consecutive 0's after an odd number of consecutive l's~\citep{tomita:cogsci82}. Initializing $\tau=1$, we train \st (with $k=10$ states) to learn a \st model to represent this grammar and then extract DFAs (see Appendix \ref{sec:dfa_extraction} for the extraction algorithm). In principle, $k \geq $ \# of classes, the model learns to select a finite set of (meaningful) states to represent a language, as shown in Appendix Figure 8. \st is able to learn that the underlying language is deterministic, learning the temperature parameter $\tau$ accordingly and the extraction produces the correct underlying DFA. The details are discussed in the appendix. 

\textbf{Stochastic Languages}
We explore whether it is possible to recover a PA from a \st model trained on the data generated by a given PA. While probabilistic deterministic finite automata (PDFAs) have been extracted previously~\citep{WeissETAL:19}, to the best of our knowledge, this is the first work to directly learn a PA, which is more expressive than a PDFA \citep{denis2004learning}, by extraction from a trained RNN.
We generate training data for two stochastic languages defined by the PAs shown in Figure \ref{fig:pfsm_ndfa} (a) and (c). Using this data, we train a \st with a GRU with $k=4$ states and we directly use the Gumbel softmax distribution to approximate the probability transitions of the underlying PA (see  Appendix \ref{sec:pa} for more details and the extraction algorithm).
Figures \ref{fig:pfsm_ndfa} (b, d) depict the extracted PAs. For both stochastic languages the extracted PAs indicate that \st is able to learn the probabilistic dynamics of the ground-truth PAs.

\begin{table}[t!]
\vspace{-3mm}
\centering
\footnotesize
\begin{tabular}{l|lll|lll}
\hline
Dataset & \multicolumn{3}{c|}{\textbf{BIH}} & \multicolumn{3}{c}{\textbf{IMDB}} \\
Metrics & PE & ECE & MCE & PE & ECE & MCE \\ \hline \hline
LSTM & \textbf{1.40} & 0.78 & 35.51 & \textbf{10.42} & 3.64 & 11.24 \\
Ensembles & \underline{1.51}$\pm1e^{-3}$ & 0.72$\pm0.10$ & 31.49$\pm8.52$ & \underline{10.56}$\pm3e^{-3}$ & 3.45$\pm1.47$ & 12.16$\pm5.68$ \\
BBB & 4.69$\pm2e^{-3}$ & 0.54$\pm0.11$ & \textbf{12.44}$\pm7.65$ & 10.84$\pm2e^{-4}$ &\underline{2.10}$\pm0.02$ & \underline{6.15}$\pm0.25$ \\
VD & \underline{1.51}$\pm3e^{-4}$ & 0.80$\pm0.03$ & 24.71$\pm16.70$ & \underline{10.56}$\pm6e^{-4}$ & 3.41$\pm0.07$ & 14.08$\pm0.87$ \\
\st $k=5/2$ \hspace{-2mm}  & 2.12$\pm5e^{-4}$ & \underline{0.45}$\pm0.03$ & 23.11$\pm12.76$ & 10.95$\pm5e^{-4}$ & \textbf{0.89}$\pm0.05$ & \textbf{3.70}$\pm0.63$ \\
\st $k=10$  &2.11$\pm5e^{-4}$&\textbf{0.40}$\pm0.05$&\underline{21.73}$\pm16.15$&11.16$\pm7e^{-4}$&3.38$\pm0.05$&9.09$\pm0.74$\\ 

\hline
\end{tabular}
\caption{\label{tab:bih_imdb}Predictive Error (PE) and calibration errors (ECE, MCE) for the datasets BIH and IMDB (lower is better for all metrics). \st offers the best and reliable trade-off between predictive error and calibration errors. Furthermore, it does not require more parameters as BBB (double) or Deep Ensemble (order of magnitude more) nor does a hyperparameter has to be tuned as in VD. Stochastic predictions are averaged across 10 independent runs and their variance is reported. Best and second best results are marked in bold and underlined  (PE: bold models are significantly different at level $p \le 0.005$). An ablation experiment with post-hoc temperature scaling is in Appendix~\ref{sec:temperature_scaling}.}
\end{table}

\begin{figure*}[t!]
	\vspace{-3mm}
	\centering
	\subfloat[LSTM (error 1.32\%)]{{\includegraphics[width=0.23\textwidth]{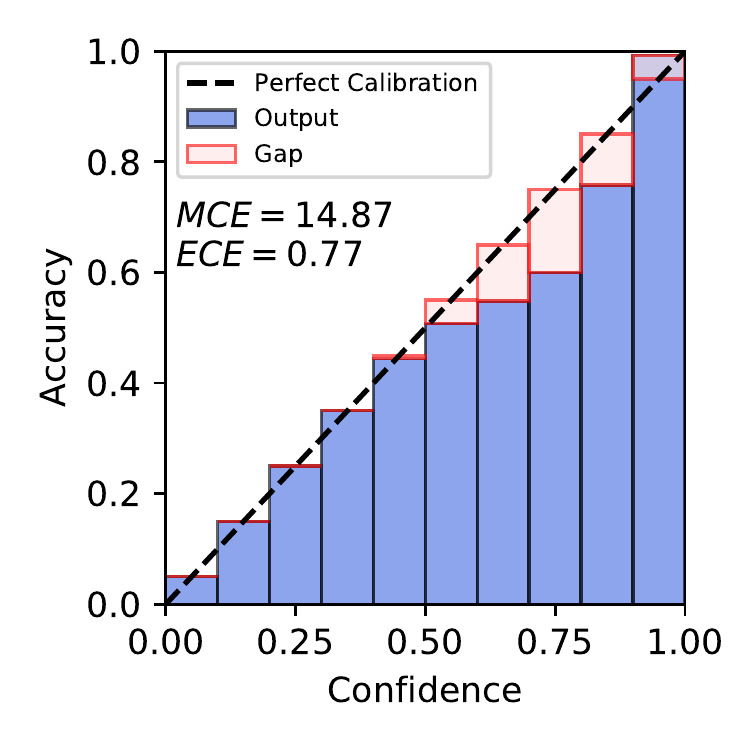} }}
	\subfloat[BBB (error 4.29 \%)]{{\includegraphics[width=0.23\textwidth]{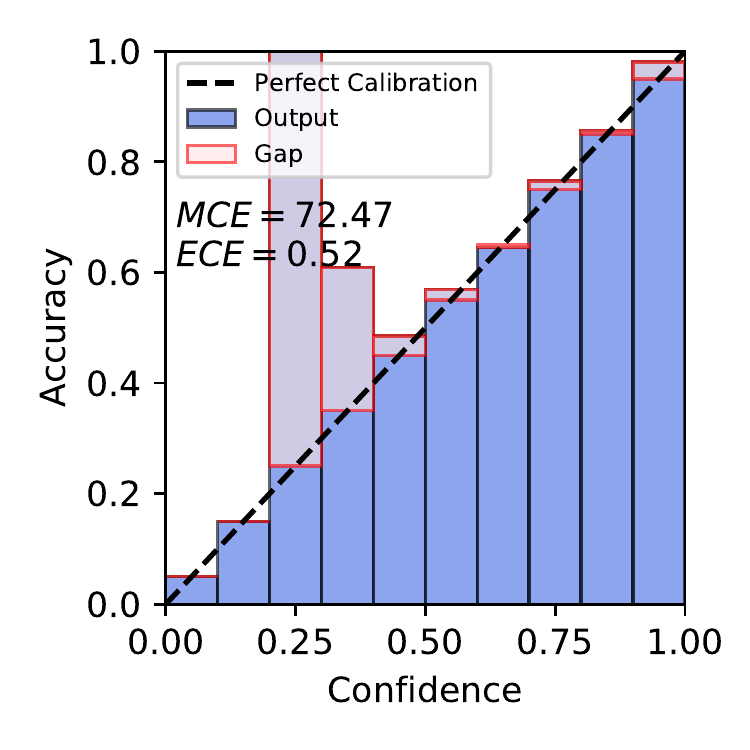} }}
	\subfloat[VD 0.05 (error 1.45\%)]{{\includegraphics[width=0.23\textwidth]{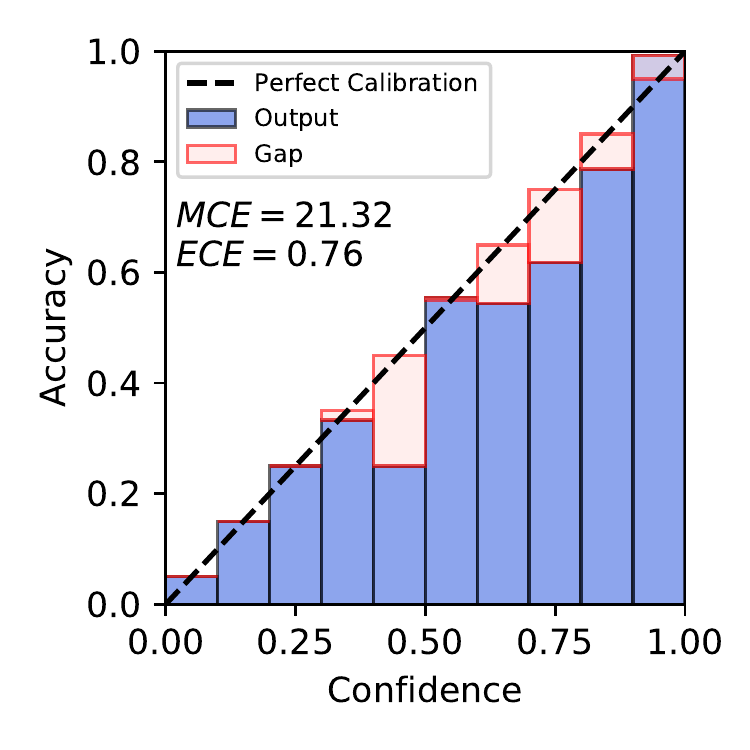} }}
	\subfloat[\st (error 2.09\%)]{{\includegraphics[width=0.23\textwidth]{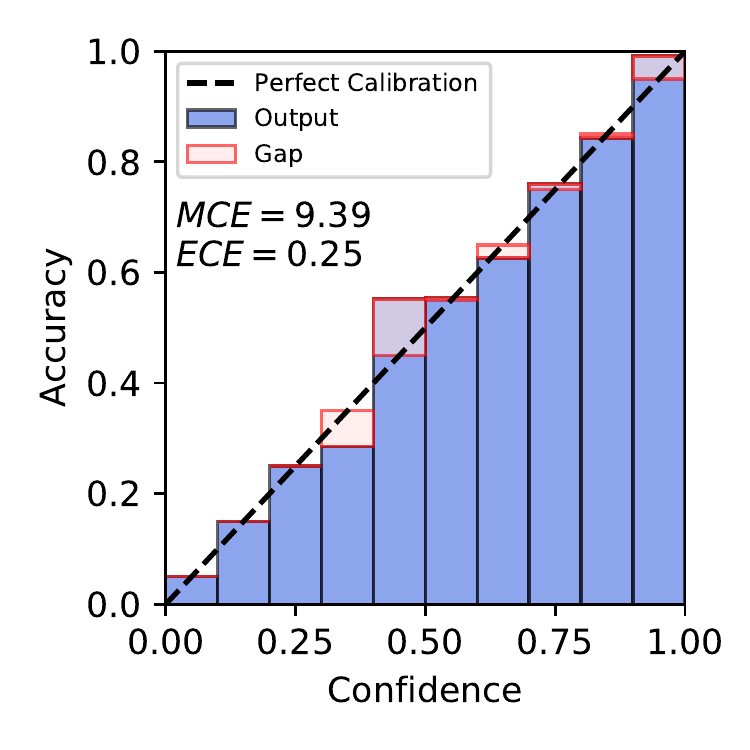} }}
	\caption{\label{fig:bih_calibration} Calibration plots (for one run) on the BIH dataset (corresponding to the left half of Table \ref{tab:bih_imdb}) using $N=10$ bins. \st ($k=2$) is closest to the diagonal line, that is, perfect calibration.}
\vspace{-2mm}
\end{figure*}

\subsection{Model Calibration}\label{sec:class}

We evaluate \st's prediction and calibration quality on two classification tasks. The first task is heartbeat classification with 5 classes where we use the MIT-BIH arrhythmia dataset~\citep{GoldbergerETA:00,MoodyMark:01}. It consists of $48$ half-hour excerpts of electrocardiogram (ECG) recordings. To preprocess the recordings we follow \cite{KachueeETAL:18} (Part III.A). The second task is sentiment analysis where natural language text is given as input and the problem is binary sentiment classification. We use the IMDB dataset \citep{MaasETAL:11}, which contains reviews of movies that are to be classified as being positive or negative. For further dataset and hyperparameter details, please see Appendix \ref{app:class:data}.

We compute the output of each model $10$ times and report mean and variance. For the deep ensemble, we train $10$ distinct LSTM models. For VD, we tuned the best dropout rate in $\{0.05, 0.1, 0.15\}$ and for BBB we tuned $\mu=\{0, 0.01\}$ and $\rho=\{-1, -2, -3, -4\}$, choosing  the best setup by lowest predictive error achieved on validation data. For \st, we evaluate both, setting the number of states to the number of output classes $(k=5/2$, BIH and IMDB, respectively$)$ and to a fixed value $k=10$. We initialize with $\tau=1$ and use a dense layer before the softmax classification. For more details see Appendix, Table \ref{tab:hyper}.
With a perfectly calibrated model, the probability of the output equals the confidence of the model. However, many neural networks do not behave this way \cite{guo2017calibration}. To assess the extent to which a model is calibrated, we use reliability diagrams as well as Expected Calibration Error (ECE) and Maximum Calibration Error (MCE). Please see Appendix~\ref{app:class:eval} for details.
For VD, the best dropout probability on the validation set is 0.05. Lower is better for all metrics. For PE, all models are marked bold if there is no significant difference at level $p \le 0.005$ to the best model.

\textbf{Results}. 
The results are summarized in Table \ref{tab:bih_imdb}. For the BIH dataset, the vanilla LSTM achieves the smallest PE with a significant difference to all other models at $p \le 0.005$ using an approximate randomization test \citep{Noreen:1989}. It cannot, however, measure uncertainty and suffers from higher ECE and MCE. Similarly, VD exhibits a large MCE. The situation is reversed for BBB were we find a vastly higher PE, but lower MCE. In contrast, \st achieves overall good results: PE is only slightly higher than the best model (LSTM) while achieving the lowest ECE and the second lowest MCE. The calibration plots of Figure \ref{fig:bih_calibration} show that \st is well-calibrated in comparison to the other models.
For the IMDB dataset, \st has a slightly higher PE than the best models, but has the lowest ECE and MCE offering a good trade-off. The calibration plots of IMDB can be found in the Appendix, Figure \ref{fig:imdb_calibration}. 
In addition to achieving consistently competitive results across all metrics, \st has further advantages compared to the other methods. The deep ensemble doubles the number of parameters by the number of model copies used. BBB requires the doubling of parameters and a carefully chosen prior, where \st does only require a slight increase in number of parameters compared to a vanilla LSTM. VD requires the tuning of the hyperparameter for the dropout probability, which leads to a trade-off between predictive and calibration errors (see Appendix \ref{sec:app:vd}).

\begin{figure}[!t]
\vspace{-3mm}
\footnotesize
\center
\begin{tabular}{l|llllll}
\hline
Datasets                 & \multicolumn{3}{c|}{IMDB(In)/Customer(Out)} & \multicolumn{3}{c}{IMDB(In)/Movie(Out)}                                                  \\ 
Method                   & Accuracy                       & O-AUPR                       & \multicolumn{1}{l|}{O-AUROC}                      & Accuracy                       & O-AUPR                       & O-AUROC                      \\                                                \hline
\hline
LSTM (max. prob.)          & 87.9                           & 72.5                           & \multicolumn{1}{l|}{77.3}                           & 88.1                           & 66.7                           & 71.6                           \\
VD 0.8                   & \textbf{88.5} & 74.8                           & \multicolumn{1}{l|}{80.6}                           & 87.5                           & 69.3                           & 74.7                           \\
BBB $\rho=-3$               & 87.6                           & 67.4                           & \multicolumn{1}{l|}{72.0}                           & 87.6                           & 67.1                           & 71.9                           \\
ST-$\tau$ $k=10$               & 88.3                           & \textbf{80.1} & \multicolumn{1}{l|}{\textbf{84.5}} & \textbf{88.1} & \textbf{75.1} & \textbf{81.0} \\ 
\hline
\hline
VD 0.8                   & \textbf{88.5} & 67.8                           & \multicolumn{1}{l|}{76.5}                           & 87.5                           & 63.8                           & 71.8                           \\
BBB $\rho =-3$               & 87.6                           & 76.0                           & \multicolumn{1}{l|}{75.4}                           & 87.6                           & \textbf{76.8} & 75.6                           \\
ST-$\tau$ $k=100$              & 86.5                           & \textbf{78.9} & \multicolumn{1}{l|}{\textbf{82.8}} & 85.9                           & 74.0                           & \textbf{78.7} \\
ST-$\tau$ $k=10$               & 88.3                           & 65.0                           & \multicolumn{1}{l|}{76.5}                           & \textbf{88.1} & 64.1                           & 75.1                           \\
\hline
\hline
Ensembles (max.prob.)         & \textbf{88.6} & 78.9                           & \multicolumn{1}{l|}{\textbf{84.4}} & \textbf{88.3} & 74.5                           & 79.6                           \\
Ensembles (variance)          & \textbf{88.6} & \textbf{79.7} & \multicolumn{1}{l|}{84.0}                                                & \textbf{88.3} & \textbf{75.8} & 79.9                           \\
\hline               
\end{tabular}
\captionof{table}{\label{tab:ood}Results (averaging on 10 runs for VD, BBB, \st.  Ensembles are based on 10 models) of the out-of-distribution (OOD) detection with max-probability based (top), variance of max-probability based (middle) and ensembles (bottom). \st exhibits very competitive performance.}
\end{figure}

\subsection{Out-Of-Distribution Detection}
\label{subsection-ood-experiments}

We explore the ability of \st to estimate uncertainty by making it detect out-of-distribution (OOD) samples following prior work~\citep{hendrycks2016baseline}. The in-distribution dataset is IMDB and we use two OOD datasets: the Customer Review test dataset~\citep{hu2004mining} and the Movie review test dataset~\citep{pang-etal-2002-thumbs}, which consist of, respectively, 500 and 1,000 samples. 
As in ~\cite{hendrycks2016baseline}, we use the evaluation metrics AUROC and AUPR
. Additionally, we report the accuracy on in-distribution samples. For VD we select the dropout rate from the values $\{0.05, 0.1, 0.2\}$ and for BBB we select the best $\mu=\{0, 0.01\}$ and $\rho=\{-1, -2, -3, -4\}$, based on best AUROC and AUPR. For \st we used $c=10$ and $c=100$. Beside using the maximum probability of the softmax (MP) as baseline~\citep{hendrycks2016baseline}, we also consider the variance of the maximum probability (Var-MP) across $10$ runs. The number of in-domain samples is set to be the same as the number of out-of-domain samples from IMDB~\citep{MaasETAL:11}. Hence, a random baseline should achieve 50\% AUROC and AUPR. 

\textbf{Results}. 
Table \ref{tab:ood} lists the results. \st and deep ensembles are the best methods in terms of OOD detection and achieve better results for both MP and Var-MP. The MP results for \st are among the best showing that the proposed method is able to estimate out-of-distribution uncertainty. 
We consider these results encouraging especially considering that we only tuned the number of learnable finite states $c$ in \st. Interestingly, a larger number of states improves the variance-based out-of-distribution detection of \st. In summary, \st is highly competitive in the OOD task.

%% file: sec-conclusion.tex
\section{Discussion and Conclusion}
We proposed \st, a novel method to model uncertainty in recurrent neural networks. \st achieves competitive results relative to other strong baselines (VD, BBB, Deep Ensembles), while circumventing some of their disadvantages, e.g., extensive hyperparameters tuning and doubled number of parameters. \st provides a novel mechanism to capture the uncertainty from (sequential) data over time steps. The key characteristic which distinguishes \st from baseline methods is its ability to model discrete and stochastic state transitions using modern gradient estimators at each time step.

%% file: sec-appendix.tex
\section{Deterministic \& Probabilistic Automata Extraction}
\label{app:dfa_pa}

For the \st  models we use an embedding layer and an LSTM layer (both with 100 hidden units) and a dense layer which accepts the last hidden state output and has two output neurons (accept, reject). The training objective aims to minimize a cross-entropy loss for a binary classification problem (accept or reject). 

\subsection{Definition of Deterministic Finite Automata}

A Deterministic Finite Automata (DFA) is a 5-tuple $(\calQ, \Sigma, \delta, q_0, F)$ consisting of a finite set of states $\calQ$; a finite set of input tokens $\Sigma$ (called the input alphabet); a transition function $\delta : \calQ \times \Sigma   \rightarrow \calQ$; a start state $q_0$; and a set of accept states $F \subseteq \calQ$.
Given a DFA and an input, it is possible to follow how an accept or reject state is reached. DFAs can be extracted from RNNs in order to offer insights into the workings of the RNN, making it more interpretable. \srrns \citep{WangNiepert:19} extract DFAs from RNNs by counting the number of transitions that have occurred between a state and its subsequent states, given a certain input~\citep{Schellhammer:1998}. However, this extraction method is deterministic and cannot give any uncertainty estimates for the extracted DFA. By adding stochasticity using the Gumbel softmax distribution, we can additionally offer uncertainty measures for the state transitions. 

\subsection{Extracting Automata for Regular Languages}

\begin{figure}[ht]
	\centering
	\subfloat[Temperature $\tau$ over iterations]{{\includegraphics[width=0.7\textwidth]{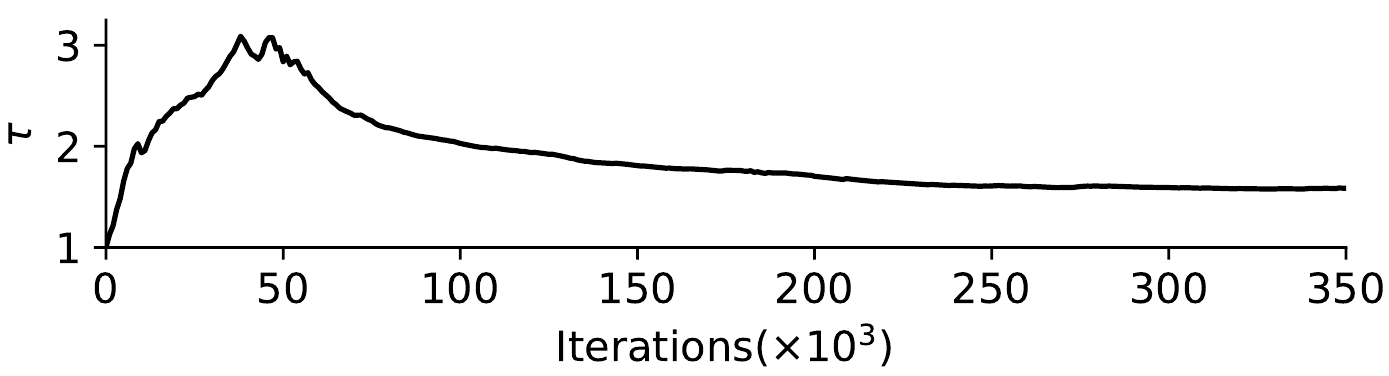} }}\\
	\subfloat[Extracetd DFA after 50$\times 10^3$ iterations, $\tau=$2.98.]{{\includegraphics[width=0.7\textwidth]{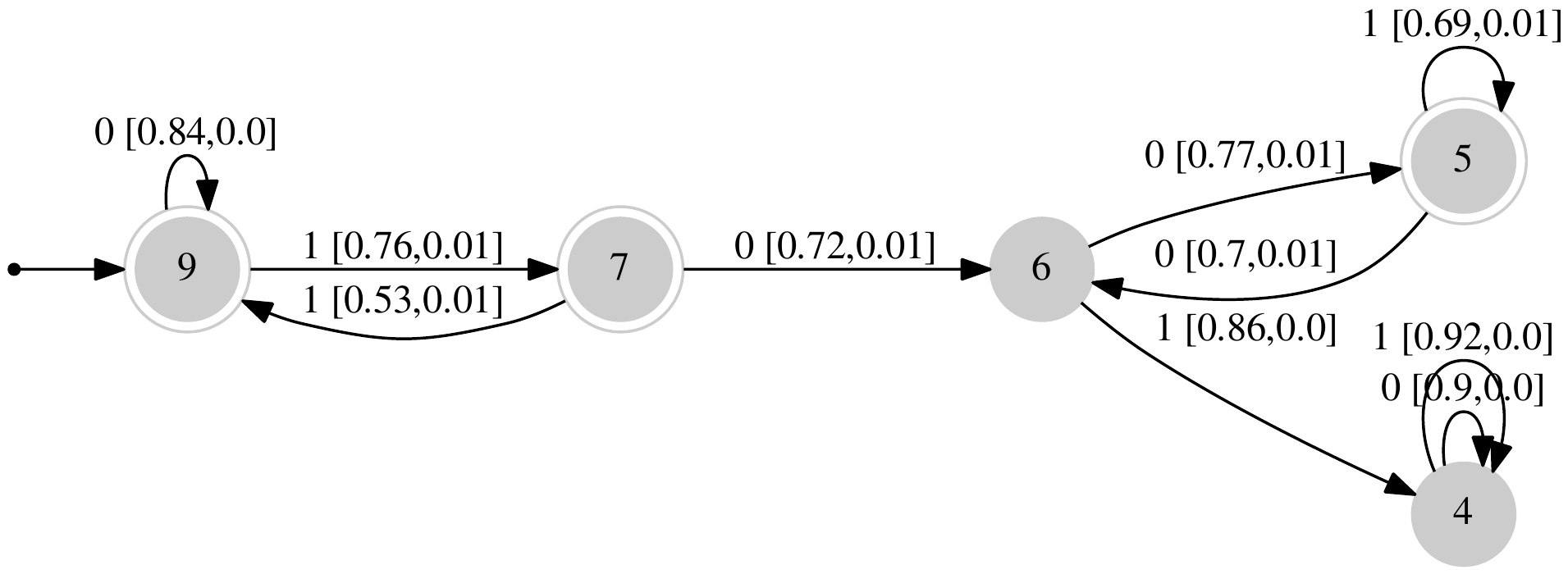} }}
	
	\subfloat[Extracetd DFA after 250$\times 10^3$ iterations, $\tau=$1.61]{{\includegraphics[width=0.7\textwidth]{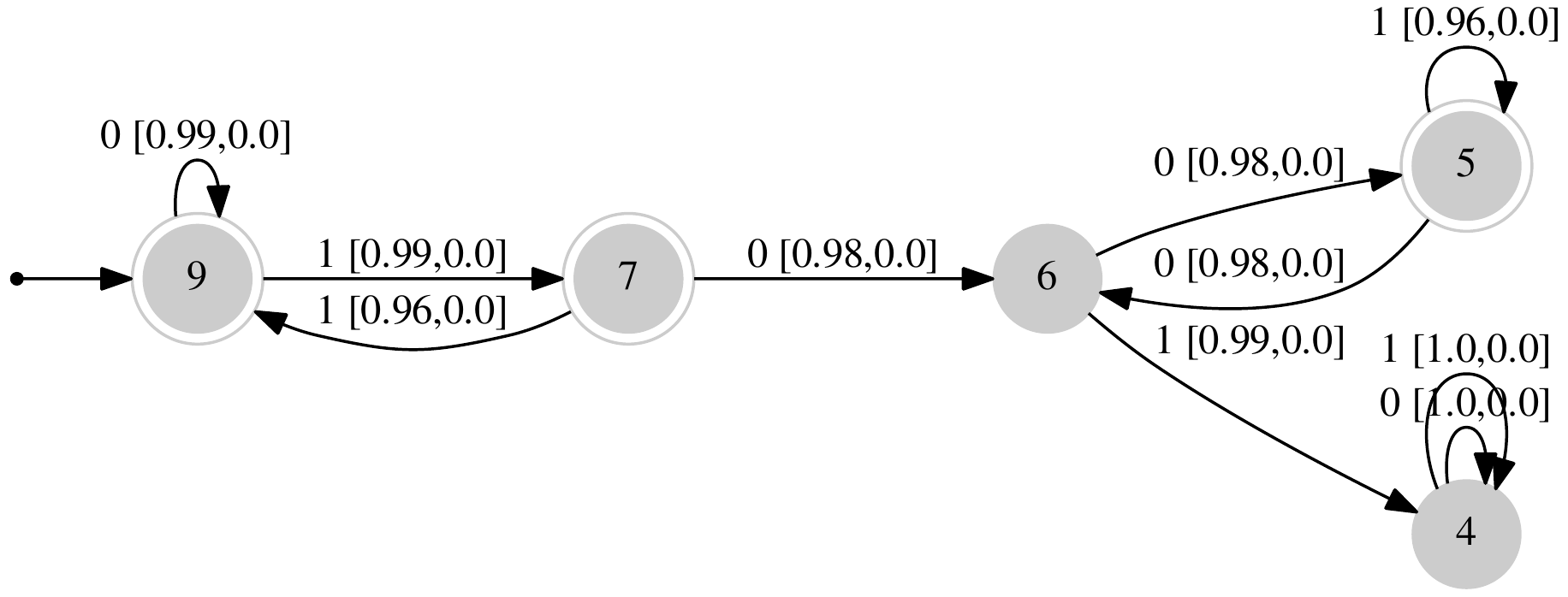} }}
	\caption{\label{fig:pfsm_t3} (a) As training progresses the  learned temperature $\tau$ decreases. This indicates that the model has reduced its epistemic uncertainty. (b, c) The extracted DFAs at different iterations on the Tomita grammar $3$ with input symbol and, in square brackets, the transition probability and uncertainty, quantified by the variance. At the earlier stage in training (b), the probabilities are still far away from being deterministic, which is also indicated by the higher value of $\tau$ and the non-zero variance. Later (c), the transition probabilities are close to being deterministic, the temperature has lowered and there is no more epistemic uncertainty. States {1-3, 8, 10} are missing as the model choose not to use these states.}
\end{figure}

With this experiment, we want to explore two features of \st. First, we want to understand how $\tau$ changes as training progresses (see Figure \ref{fig:pfsm_t3} (a)). At the beginning of training, $\tau$ first increases, allowing the model to explore the state transitions and  select the states which will represent the corresponding grammar (in our example, the model selects 5 out of 10 states to represent Tomita 3, see Figure \ref{fig:pfsm_t3} (b, c)). Later, $\tau$ decreases and the transition uncertainty is calibrated to have tighter bounds, becoming more deterministic). 
Second, we want to see if \st can model the uncertainty in transitions and adaptively learn to calibrate the state transition distributions. For this, we extract DFAs at two different iterations (see Figure \ref{fig:pfsm_t3} (b, c)). After 50k iterations, a correct DFA can be extracted. However, the transitions are not well calibrated. The ideal transition should be deterministic and have transition probability close to 1. For example, at state 7, for input ``\textit{1}'', only 53\% of the time the model transitions to the correct state 9. In contrast, after 250k iterations, the transitions are well calibrated and all transition are almost deterministic. At the same time, $\tau$ has lowered, indicating that the model has become more certain. 
\subsection{Definition of Probabilistic Automata}

The stochasticity in \st also allows us to extract Probabilistic Automata (PA)~\citep{rabin1963probabilistic} which have a stronger representational power than DFAs. A PA is defined as a tuple $(\calQ, \Sigma, \delta, q_0, F)$ consisting of a finite set of states $\calQ$; a finite set of input tokens $\Sigma$; a transition function $\delta : \calQ \times \Sigma   \rightarrow \calP$, where $\calP$ is the transition probability for a particular state, and $\calP(\calQ)$ denotes the power set of $\calQ$; a start state $q_0$ and a set of accept states $F \subseteq \calQ$. 

\subsection{Extracting DFAs with Uncertainty Information}
\label{sec:dfa_extraction}
Let $p(s_j|s_i, x_t)$, $i, j \in \{1, ..., k\},~~ x_t \in \Sigma$ be the probability of the transition to state $s_j$, given current state $s_i$ and current input symbol $x_t$ at a given time step $t$.  We query each training sample to model and record the transition probability for each input symbol. The DFA extraction algorithms in \citep{Schellhammer:1998,Wang:2007:nc} are based on the count of transitions. In contrast, our extraction algorithm utilizes the transition probability, as described in Algorithm \ref{alg:1}.

\begin{algorithm}[ht]
	\renewcommand{\algorithmicrequire}{\textbf{Input:}}
	\renewcommand{\algorithmicensure}{\textbf{Output:}}
	\caption{Extracting DFAs with Uncertainty Information}
	\label{alg:1}
	\begin{algorithmic}[1]
		\REQUIRE model $\mathcal{M}$, dataset $\textbf{D}$, alphabet $\Sigma$, start token $x_0$
		\ENSURE transition function $\delta$
		\STATE Initialize an empty dictionary $Z[s_j|s_i, x_t]=\O$
		\STATE Compute the probability distribution over $k$ states when input $x_0$: \\ $p_{1:k}=\mathcal{M}(x_0)$,  $s_ j=\argmax_{i \in \{1, ..., k\}} [p_{1:k}]$
		\STATE Set $s_i=s_j$, Update $Z[s_j|x_0]=\{p_j\}$
		\FOR{$\bfx = (x_1, x_2,..., x_T) \in \textbf{D}$}   
     		\FOR{$t \in [1, ..., T]$}  
		\STATE  $p_{1:k}=\mathcal{M}(s_i, x_t), s_j = \argmax_{i \in \{1, ..., k\}} [p_{1:k}]$,
		\STATE  Set $s_i=s_j$,  Update $Z[s_j|s_i, x_t]$: \\
			 		\STATE $Z[s_j|s_i, x_t] = Z[s_j|s_i, x_t] \cup \{p_j\}$, 
		\ENDFOR
		\ENDFOR
		\STATE  Compute transition function $\delta$, transition probability mean $p_u$ and variance $p_{var}$:
		\FOR{$i, j \in \{1, ..., k\} \mbox{ and } x_t \in \Sigma$}   
		\STATE $p_u=mean(Z[(s_j|s_i, x_t)])$
		\STATE $p_{var}=var(Z[(s_j|s_i, x_t)])$
		\STATE $ \delta(s_j|s_i, x_t) = \argmax_{j \in \{1, ..., k\}} p_u  $
		\ENDFOR
	\end{algorithmic}  
\end{algorithm}

\subsection{Extracting Probabilistic Automata}
\label{sec:pa}
We first define two probabilistic automata as shown in Figure \ref{fig:pfsm_ndfa}(a)(c) to generate samples from it\footnote{We generate samples without combining identical samples. For instance, consider a sequence drawn from PA-2 which has probability 0.7 to be rejected and probability  0.3 to be accepted. In this case, we generate 10 training samples with the sequence, 7 of which are labeled ``0" (reject) and 3 of which are labeled ``1" (accept) in expectation.}.  We generated 10,170 samples for stochastic language 1 (abbreviated SL-1) with PA-1 (Figure \ref{fig:pfsm_ndfa}(a)) and sample length $l \in [1,9]$. For SL-2 with PA-2, we generated 20,460 samples with sample length $l \in [1,10]$. The learning procedure is described in Algorithm~\ref{alg:pa}. We use the Gumbel softmax distribution in \st to approximate the probability distribution of next state $p(s_j|s_i, x_t )$. To achieve this goal, we set the number of states $k$ to an even number and force the first half of the state to ``reject’’ state and the second half of states to be ``accept’’. This allows us to ensure that the model models both ``reject’’ and ``accept’’ with the same number of states.

\begin{algorithm}[ht]
	\renewcommand{\algorithmicrequire}{\textbf{Input:}}
	\renewcommand{\algorithmicensure}{\textbf{Output:}}
	\caption{Extracting PAs with \st}
	\label{alg:pa}
	\begin{algorithmic}
		\REQUIRE dataset $\textbf{D}$
		\ENSURE network loss $L$
		\FOR{$\bfx, \bfy \in \textbf{D}$} 
		\STATE initialize $h_0$ with the $1^{st}$ state $s_1$, $s_1 \in \mathbf{S}$=$\{s_{1:k}\}$
		\STATE  $h_0=s_1, \hat{\bfy}=0$
     			\FOR{$x_t \in \bfx, t \in [1, ..., T]$}  
			\STATE  $z_t=sigma(W_zx_t+U_zh_{t-1})$
			\STATE  $r_t=sigma(W_rx_t+U_rh_{t-1})$
			\STATE  $g_t=tahn(W_gx_t+U_gh_{t-1})$
			\STATE  $u_t=z_t\odot h_{t-1}+(1-z_t)\odot g_t$
			\STATE  $p_t=\mathbf{S} u_t^\top$
			\STATE  $\hat{p_t}\sim $\textsc{Gumbel}$(p_t, \tau)$
			\STATE  $h_t=\hat{p_t}^\top \mathbf{S}$
			\STATE  $\hat{\bfy}=\hat{p_t}$
			\ENDFOR
		\IF {$\bfy=0$}
			\STATE $\bfy =[\underbrace{0,0,..0,}_{k/2} \underbrace{1,1,..1}_{k/2}] $
		\ELSE
			\STATE $\bfy =[\underbrace{1,1,..1,}_{k/2} \underbrace{0,0,..0}_{k/2}] $
		\ENDIF
		\STATE  Compute cross-entropy loss: 
		\STATE $L=$\textsc{cross-entropy}$(\bfy, \hat{\bfy})$
		\ENDFOR
	\end{algorithmic}  
\end{algorithm}
\begin{figure}[ht]
	\centering
    \subfloat[Truth PA-2]{{\includegraphics[width=0.265\textwidth]{figures/g2_truth.pdf} }}
    \subfloat[Extracted PA-2]{{\includegraphics[width=0.275\textwidth]{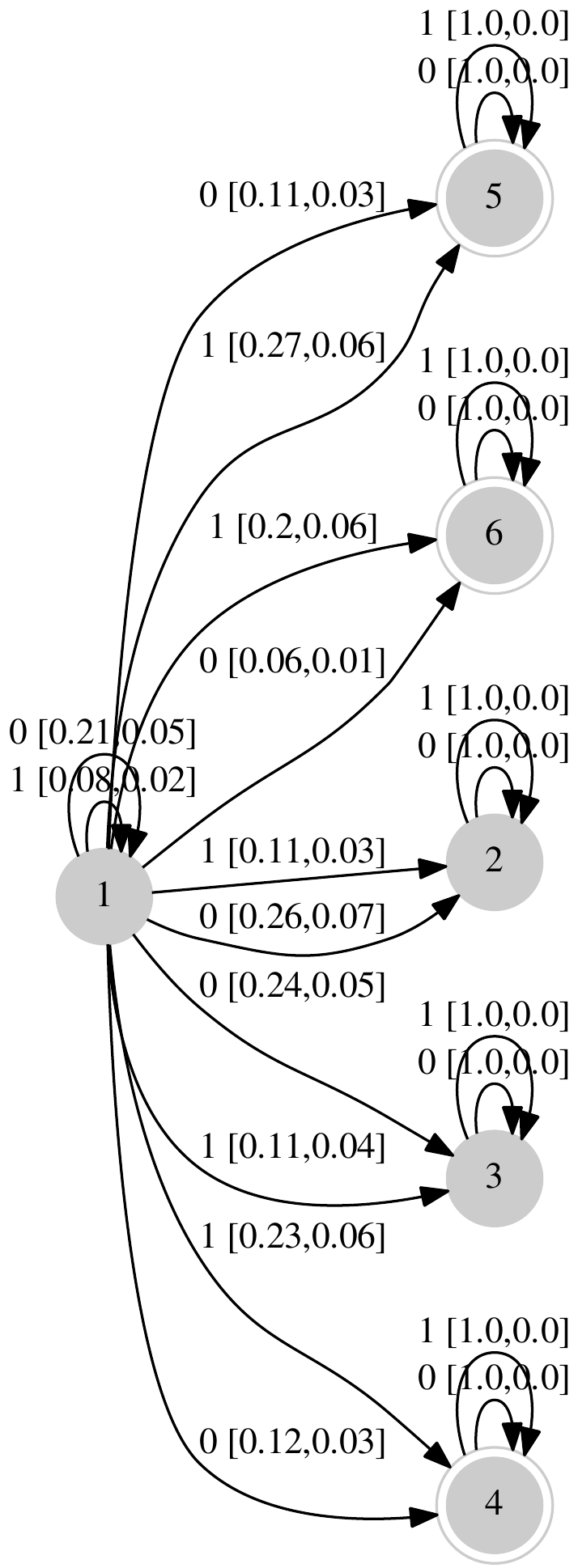} }}
	\subfloat[Minimized PA-2]{{\includegraphics[width=0.215\textwidth]{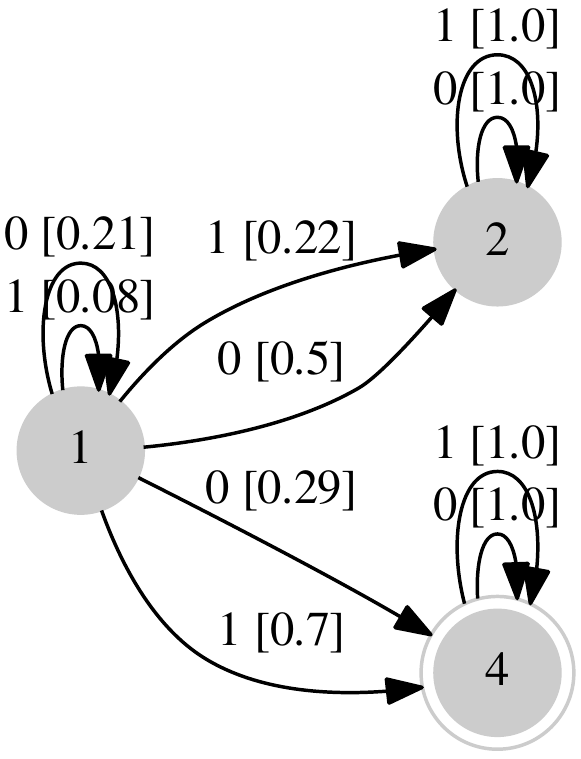} }}
	\caption{\label{fig:pfsm_ndfa2} True PA (a) used for generating training samples and the respective extracted PAs (b, c) from the trained \st models with $k=6$ states. We merged reject and accept nodes in (b) to retain one reject (start state 1 is not merged) and accept node in (c). For a given state and input symbol, the probability of accept and reject for (a) and (c) are nearly equivalent. For example, at state $s_A$ in (a), input ``0", $p(accept| s_A, ``0")=0.3\approx0.29=p(accept|s_1, ``0")$ and $p(accept| s_A, ``0")=0.7 \approx 0.71=0.5+0.21=p(accept|s_1, ``0")$.}
\end{figure} 

In the main part of the paper we report results when setting \st to have $k=4$ states for SL=2. Here, we additionally present the results for $k=6$ in Figure \ref{fig:pfsm_ndfa2}, which yields comparable results, showing that \st is not overly sensitive to the number of states. Is is however, helpful to have the same number of accept and reject states.

\clearpage
\section{Model Calibration}

We address the problem of supervised multi-class sequence classification with recurrent neural networks. We follow the definitions and evaluation metrics as in \citep{guo2017calibration}. Given input $X\in \mathcal{X}$ and ground-truth label $Y \in \mathcal{Y}$, a probabilistic classifier $m(X)=(\hat{Y}, \hat{P})$. The $\hat{Y}=\{ \hat{y}_1,...,\hat{y}_k\}, \hat{P}=\{\hat{p}_1,...,\hat{p}_k\}$ present the predicted class label and confidence (the probability of correctness) over $k$ classes and $\sum_{i=1}^k \hat{p}_i=1$. 

\textbf{\textit{Definition of Calibration}} A model is perfectly calibrated if the confidence estimation equals the true probability, that is, $\mathbb{P}(\hat{Y}=Y|\hat{P}=p)=p$, $p \in [0,1]$.

\subsection{Evaluation of Calibration}\label{app:class:eval}

\textbf{Reliability Diagrams} \citep{DeGrootFienberg:83,Niculescu-MizilETAL:05} visualise whether a model is over- or under-confident by grouping predictions into bins according to their prediction probability. The predictions are grouped into $N$ interval bins (each of of size $1/N$) and the accuracy of samples $y_i$ wrt. to the ground truth label $\hat{y}_i$ in each bin $b_n$ is computed as:
\begin{equation}
\textrm{acc}(b_n)=\frac{1}{|b_n|}\sum_{i}^{I}\bm{1}(\hat{y}_i=y_i),
\end{equation}
where $i$ indexes all examples that fall into bin $b_n$.
Let $\hat{p}_i$ be the probability for sample $y_i$, then average confidence is defined as
\begin{equation}
\textrm{conf}(b_n)=\frac{1}{|b_n|}\sum_{i}^{I}\hat{p}_i.
\end{equation}
A model is perfectly calibrated if $\textrm{acc}(b_n)=\textrm{conf}(b_n), \forall n$ and in a diagram the bins would follow the identity function. Any derivation from this represents miscalibration. 

Based on the accuracy and confidence measures, two calibration error metrics have been introduced \citep{NaeiniETAL:15}.

\textbf{Expected Calibration Error (ECE)}. Besides the reliability diagrams, ECE is a convenient tool to have scalar summary statistic of calibration. It computes the difference between model accuracy and confidence as a weighted average across bins, 
\begin{equation}\label{eq:ece}
\textsc{ECE} = \sum_{n=1}^{N} \frac{|b_n|}{m} |\textrm{acc}(b_n) - \textrm{conf}(b_n)|,
\end{equation}
where $m$ is the total number of samples.

\textbf{Maximum Calibration Error (MCE)} is particularly important in high-risk applications where reliable confidence measures are absolutely necessary. It measures the worst-case deviation between accuracy and confidence,
\begin{equation}\label{eq:mce}
\textsc{MCE} = max_{n\in \{1, \dots, N\}} | \textrm{acc}(b_n) - \textrm{conf}(b_n)|.
\end{equation}

For a perfectly calibrated classifier, the ideal ECE and MCE both equal to 0. 
\begin{figure}[!htb]
	\vspace{-5mm}
	\centering
	\subfloat[LSTM (error 10.90\%)]{{\includegraphics[width=0.25\textwidth]{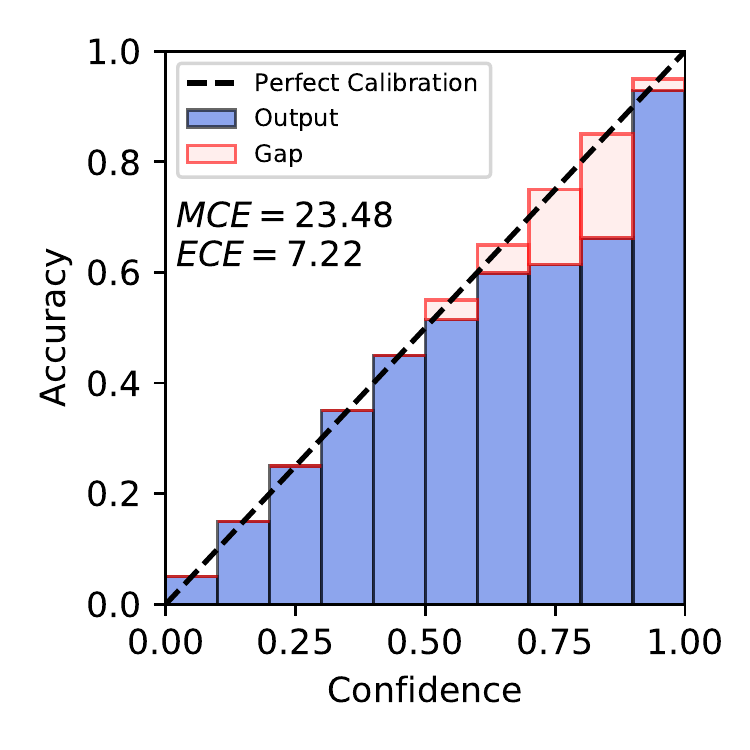} }}
	\subfloat[BBB (error 11.17\%)]{{\includegraphics[width=0.25\textwidth]{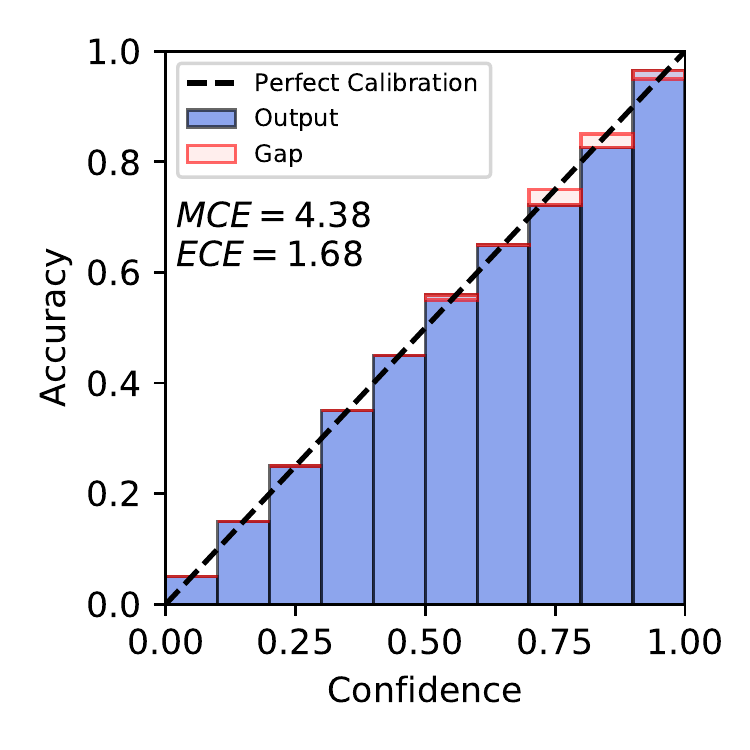} }}
	\subfloat[VD 0.1 (error 10.80\%)]{{\includegraphics[width=0.25\textwidth]{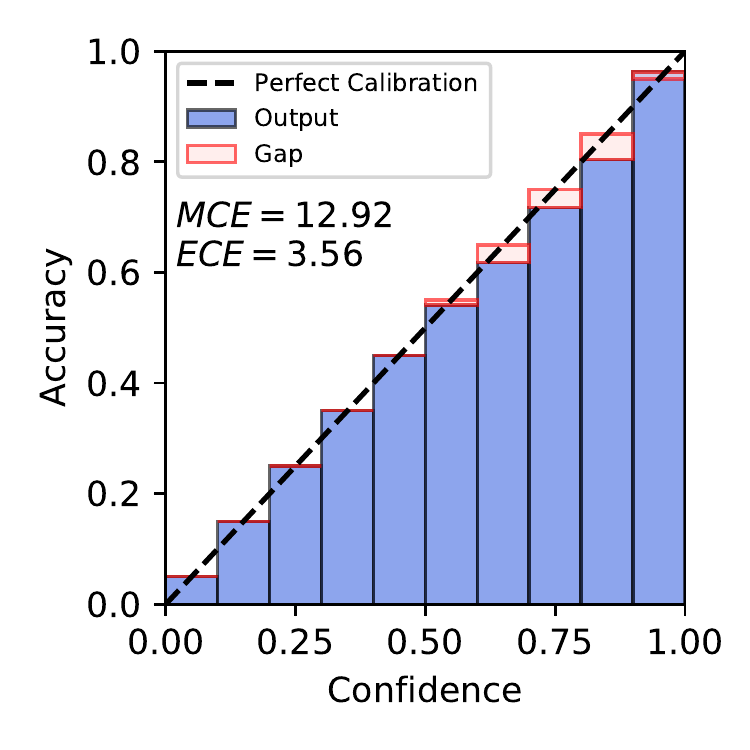} }}
	\subfloat[\st (error 10.89\%)]{{\includegraphics[width=0.25\textwidth]{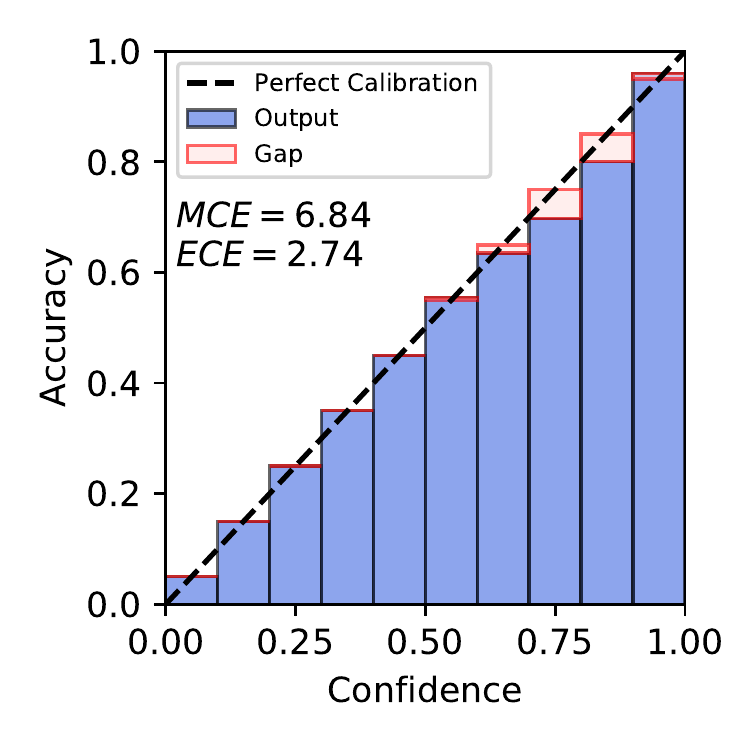} }}
	\caption{\label{fig:imdb_calibration} Calibration plots for the IMDB dataset (corresponding to the right half of Table \ref{tab:bih_imdb}) using $N=10$ bins. BBB, VD and \st are all quite well calibrated on this data set. The first run is displayed. Best viewed in colour.}
\end{figure}

\begin{table}[h]
	\begin{center}
		\begin{tabular}{lcc}
			\hline
			Dataset&IMDB&BIH\\
			\hline
			Train&23$k$&78$k$\\
			Validation&2$k$&8$k$\\
			Test&25$k$&21$k$\\
			Max Length&2,956&187\\
			\# Classes&2&5\\
			Type&Language&ECG\\
			\hline
		\end{tabular}
		\caption{Overview of used datasets, if applicable, numbers are rounded down.}
		\label{tab:datasets}
	\end{center}
\end{table}

\begin{table}[h]
	\begin{center}
		\begin{tabular}{lcc}
			\hline
			Hyperparameters&IMDB&BIH\\
			\hline
			Hidden dim.&256&128\\
			Learning rate&0.001&0.001\\
			Batch size&8&256\\
			Validation rate&1$k$&1$k$\\		
			Maximum validations&20&50\\
			\st \# states&2&5\\
			BBB $\mu$&0.0&0.01\\
			BBB $\rho$ &-3&-3\\
			VD Prob.&0.1&0.05\\
    			\hline
		\end{tabular}
		\caption{Overview of the different hyperparameters for the different datasets. Validation rate indicates after how many updates validation is performed.}
		\label{tab:hyper}
	\end{center}
\end{table}

\begin{table}[t!]
	\begin{center}
		\begin{tabular}{lllll}
			\hline
			&&PE&ECE&MCE\\
			\hline
			\multirow{5}{*}{\begin{sideways}\textsc{BIH}\end{sideways}}&LSTM&\textbf{1.40}&0.30&\underline{13.02}\\
			&Ensemble&\underline{1.51}$\pm1e^{-3}$&\textbf{0.22}$\pm0.05$&21.90$\pm16.62.52$\\
			&BBB &4.69$\pm2e^{-3}$&0.36$\pm0.09$&\textbf{10.43}$\pm7.86$\\
			&VD 0.05&\underline{1.51}$\pm3e^{-4}$&\underline{0.27}$\pm0.02$&23.60$\pm18.66$\\
			&\st&2.12$\pm5e^{-4}$&0.45$\pm0.03$&15.38$\pm7.91$\\
			\hline \hline
			\multirow{5}{*}{\begin{sideways}\textsc{IMDB}\end{sideways}}&LSTM&\textbf{10.42}&1.19&5.82\\
			&Ensemble&\underline{10.56}$\pm3e^{-3}$&1.26$\pm0.56$&5.87$\pm2.40$\\
			&BBB &10.84$\pm2e^{-4}$&\textbf{0.63}$\pm0.02$&\textbf{2.69}$\pm0.33$\\
			&VD 0.1&\underline{10.56}$\pm6e^{-4}$&2.34$\pm0.01$&10.86$\pm1.33$\\
			&\st&10.95$\pm5e^{-4}$&\underline{1.00}$\pm0.04$&\underline{3.84}$\pm0.70$\\
			\hline
			
		\end{tabular}

		\caption{Same as table\ref{tab:bih_imdb} but with post-hoc temperature scaling. Among the models (Ensemble, BBB, VD) that can estimate uncertainty, \st is very competitive, i.e., the second best on both the BIH and the IMDB datsaet. LSTM is not able to provide any uncertainty information. Predictive Error (PE) and calibration errors (ECE, MCE) for the various RNNs on the datasets BIH and IMDB (lower is better for all metrics). \st offers the best and reliable trade-off between predictive error and calibration errors. Furthermore, it does not require double the parameters as BBB nor does a hyperparameter have to be tuned as in VD. Stochastic predictions are averaged across 10 independent runs and their variance is reported. For VD, we report the best dropout probability on the validation set. Best and the second best results are marked in bold and underlined (PE: bold models are significantly different at level $p \le 0.005$ to non-bold models).}
		\label{tab:bih_imdb_1}
	\end{center}
\end{table}

\begin{figure}[htb]
\vspace{-3mm}
\centering
\subfloat[BIH]{{\includegraphics[width=0.65\textwidth]{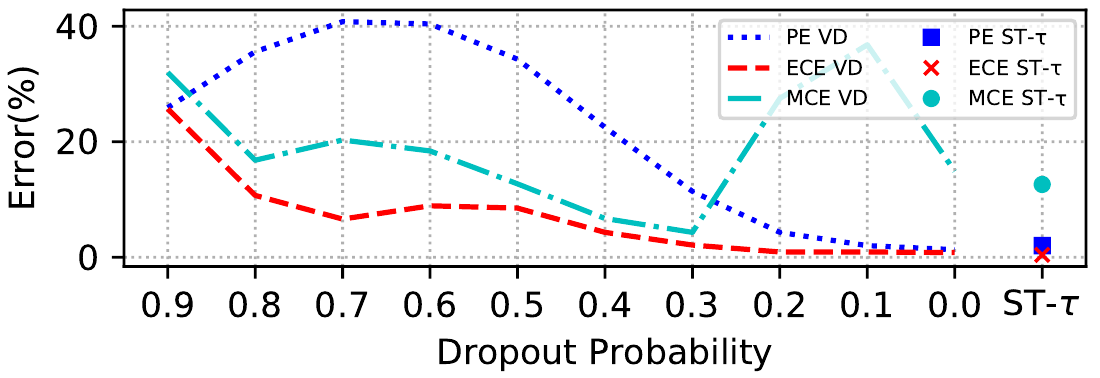} }}\\
\subfloat[IMDB]{{\includegraphics[width=0.65\textwidth]{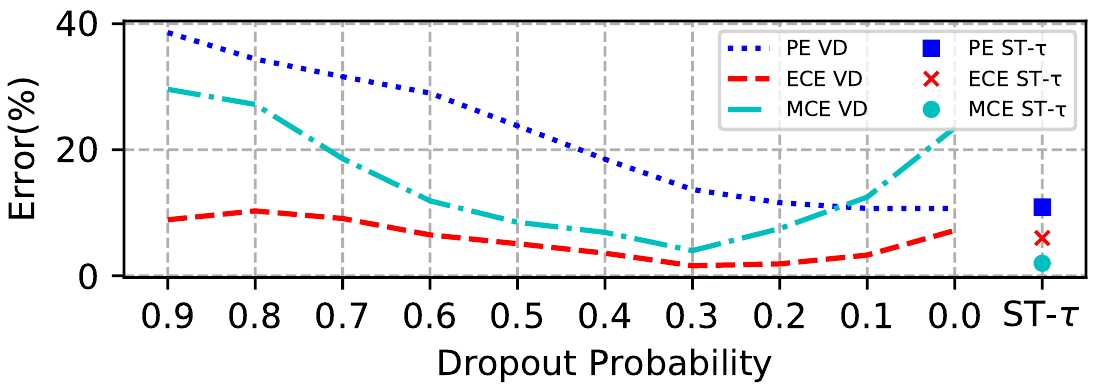} }}
\caption{\label{fig:vd} For VD we plot various dropout probabilities at prediction time and their corresponding error rates (PE, ECE, MCE). Additionally we report the results of ST-$\tau$. For VD, while lowering the probability rate decreases the predictive error (PE), it at the same time increases MCE and/or ECE. On the other hand, ST-$\tau$ combines the best of both worlds without having to tune any parameters.}
\end{figure}

\subsection{Dataset and Hyperparameter details}\label{app:class:data}
For the experiments of Section \ref{sec:class}, we provide an overview of the used datasets in Table \ref{tab:datasets} and give details on the different hyperparameters used in the experiments in Table \ref{tab:hyper}. 
On BIH, we use the training / test split of \citep{KachueeETAL:18}, however we additionally split off 10\% of the training data to use as a validation set. On IMDB, the original dataset consist of 25k training and 25k test samples, we split 2k from training set as validation dataset. The word dictionary size is limited to 5,000.

\subsection{Additional Experiments}
Here we report three groups of additional or ablation experiments: (1) All baseline method and \st with directly employing post-training temperature scaling. (2) The trade-off between predictive and calibrate performance with different dropout ratio in variational dropout (VD). (3) Additional calibration plots for the IMDB dataset. 

\subsubsection{Experiments with Temperature Scaling}
\label{sec:temperature_scaling}
For classification calibration experiments, the post-hoc temperature scaling can also be used to calibrate models. However, please note, post-hoc temperature scaling can not be used when we need to calibrate a model during training stages, for example, the tasks like DFA or PA extraction, reinforcement learning tasks. 

Table \ref{tab:bih_imdb_1} reports the results with temperature scaling where temperature is tuned on valid set. Among the models (Ensemble, BBB, VD) that can estimate uncertainty, \st is very competitive, for example, the second best on both the BIH and the IMDB datsaet. While LSTM can achieve better predictive performance and sometimes better calibration performance, LSTM is not able to provide any uncertainty information.     

\subsubsection{Experiments with Different Dropout Rate}\label{sec:app:vd}
To gain a better understanding of how crucial the hyperparameter of VD is, we investigate the effect of the dropout probability during prediction with VD. We perform an experiment where we vary the dropout probability in the range of $[0.0, 0.9]$ with increments of $0.1$. In Figures \ref{fig:vd} (a, b) we plot the result for BIH and IMDB, respectively, reporting PE, ECE and MCE for the various VD settings as well as \st. 

On both datasets, for VD, the point of lowest PE does not necessarily coincide with the points of lowest MCE and/or ECE. For example on BIH, VD achieves the lowest PE when dropout is switched off (0.0), but then uncertainty cannot be measured. On the other hand, choosing the next lowest PE results in a high MCE. In contrast, \st directly achieves good results for all three metrics. Similarly, on IMDB, at the point where VD has the lowest PE is also has highest MCE. In conclusion, VD requires careful tuning which comes with choosing a trade-off between the different metrics, while \st achieves good results directly, without any tuning.

\subsubsection{Calibration Plots for the IMDB Dataset}

Figure \ref{fig:imdb_calibration} shows the calibration plots for IMDB. For the binary classification task BBB achieves the best calibration performance and VD achieves the best predictive performance. It should be noted that \st achieves the best trade-off between predictive and calibration performance without doubling parameters (for BBB) and without tuning dropout rate for (VD). 

\begin{figure}[!t]
	\centering
	\includegraphics[width=.9\textwidth]{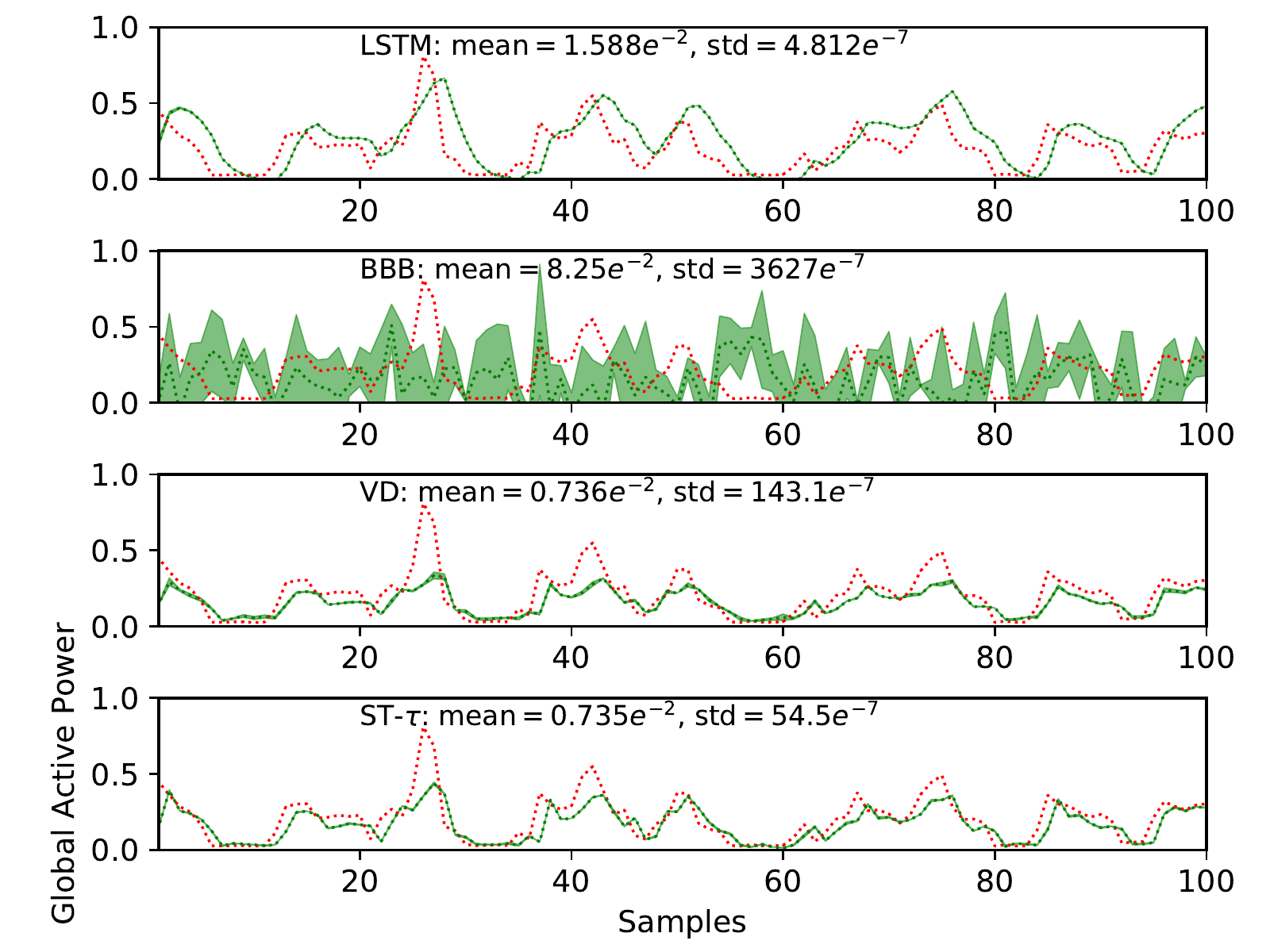}
	\caption{\label{regression} The test mean squared error (MSE) of different methods on predicting power consumption over time steps (x axis). The results are based on averaging 3 runs. The shaded areas present standard deviation, the first 100 samples are displayed. \st and VD provide the best predictive performance, while \st exhibits tighter uncertainty bounds.}
\end{figure}

\section{Regression}\label{sec:reg}
Calibration plays an important role in probabilistic forecasting. We consider a time-series forecasting regression task using the individual household electric power consumption dataset.\footnote{\url{http://archive.ics.uci.edu/ml/datasets/Individual+household+electric+power+consumption} The data is preprocessed by following \url{https://www.kaggle.com/amirrezaeian/time-series-data-analysis-using-lstm-tutorial}} The goal of the task is to predict the global active power at the current time (t) given the measurement and other features of the previous time step. The dataset was sampled at the time step of an hour (the original data are given in minutes), which leads to 34,588 samples. We split it 25,000/2,000/7,588 for training/validation/test. One LSTM layer with 100 hidden units is used for the baselines (LSTM, BBB and VD) and \st (the number of states is set to 10). The evaluation metric is mean squared error (MSE) and we use the model with lowest MSE on the validation dataset at test stage. 

Figure \ref{regression} presents the performance. LSTM, VD and \st perform very well at this task, achieving MSE close to zero. BBB performs worse than the other methods. For uncertainty estimation, BBB, VD and \st are able to provide uncertainty information alongside the predictive score. BBB gives high uncertainty, while VD and \st are more confident in their prediction, with \st offering the tightest uncertainty bounds.

\section{Reinforcement Learning}\label{sec:rl} 

\begin{figure*}
\centering
 \includegraphics[width=0.3\textwidth]{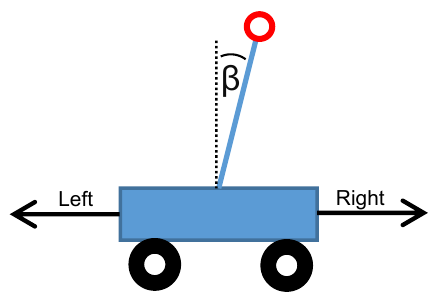}
\captionof{figure}{\label{fig:catepole}In the cartpole task the goal is to balance the pole upright by moving the cart left or right at each time step ($\beta$ is the angle).}
\end{figure*}

We explore the ability of \st to learn the exploration-exploitation trade-off in reinforcement learning. To demonstrate this empirically, we evaluate \st in the continuous control environment \textsc{cartpole} (Figure \ref{fig:catepole}) of OpenAI Gym~\citep{1606.01540}.
\begin{figure*}
	\vspace{-5mm}
	\centering
	\subfloat[MDP, sampling]{{\includegraphics[width=0.245\textwidth]{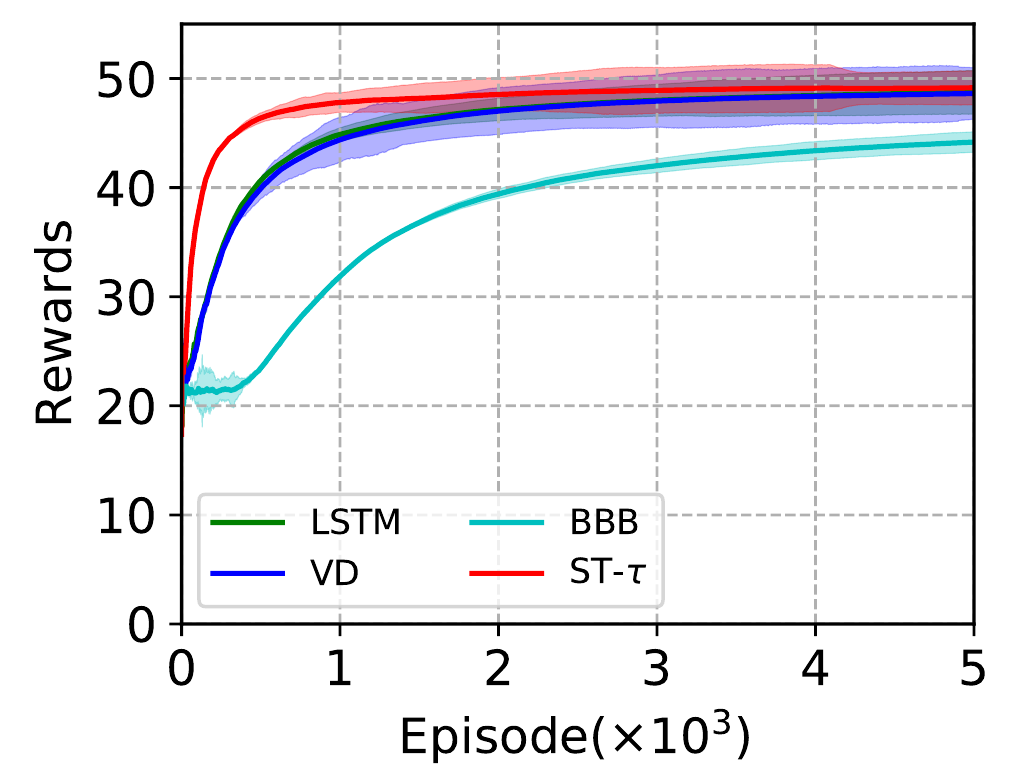} }}
	\subfloat[MDP, greedy]{{\includegraphics[width=0.245\textwidth]{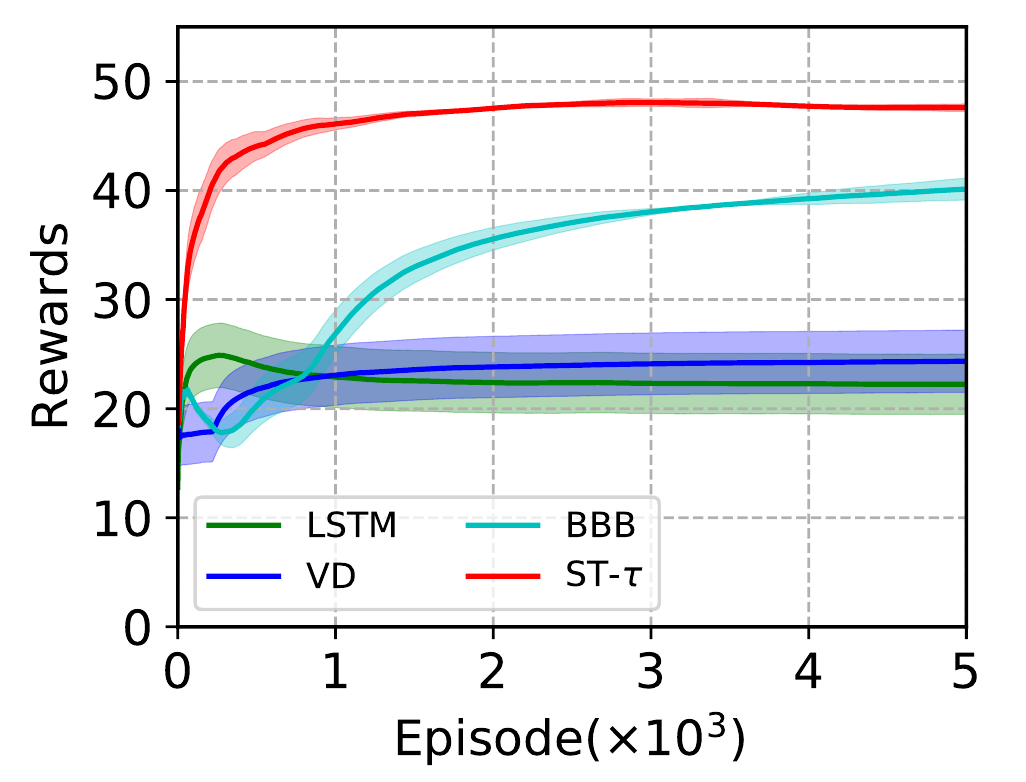} }}
	\subfloat[POMDP, sampling]{{\includegraphics[width=0.245\textwidth]{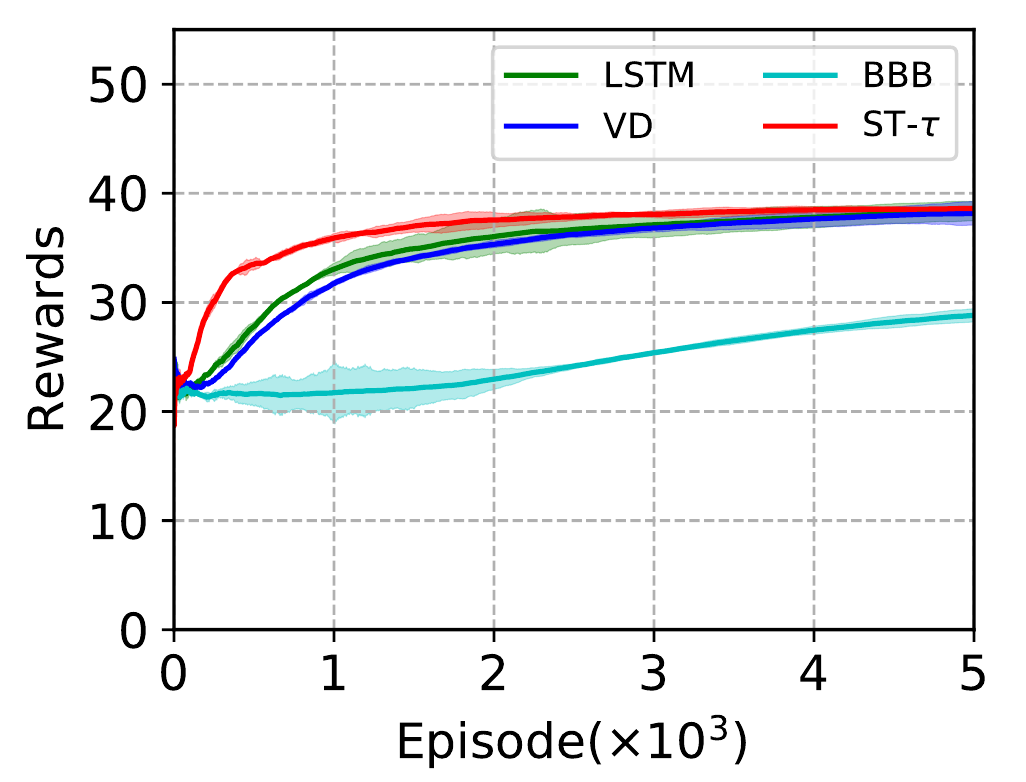} }}
	\subfloat[POMDP, greedy]{{\includegraphics[width=0.245\textwidth]{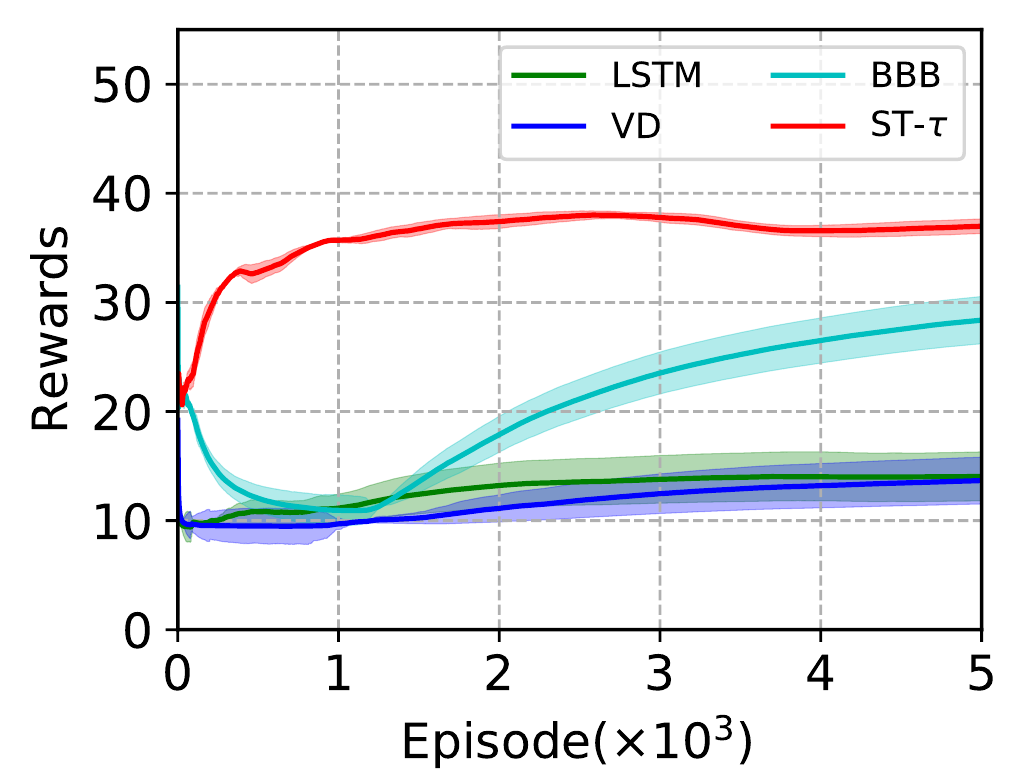} }}
	\caption{\label{fig:rl} Average cumulative reward and standard deviation (in log scale, the shade areas) over time steps (episode) on the Cartpole task. Results are averaged over 5 randomly initialized runs. In all cases \st achieves a higher averaged cumulative reward given lower sample complexity.}
\vspace{-2mm}
\end{figure*} 
The objective of the task is to train an agent to balance a pole for 1 second (which equals 50 time steps) given environment observations $O$. To keep the pole in balance, the agent has to move the pole either left or right, that is, the possible actions are $\bm{a} = \{$\textsc{left}, \textsc{right}$\}$. If the chosen action keeps the pole in balance, the agent receives a reward of $1$ and continues; otherwise a reward of $0$ is given and the run stops.

The environment can be formulated as a reinforcement learning problem ($S, A, P, R$) where $S$ is a set of states. Each state $s= \{x, \hat{x}, \beta, \hat{\beta}\} \in S$ consists of $x$: cart position,  $\hat{x}$: cart velocity, $\beta$: angle position, and $\hat{\beta}$: angle velocity. $A$ is the set of actions, $P$ the state transition probability, and $R$ the reward.
We consider two different setups. In the first setup, the \textsc{cartpole} environment is fully observable Markov Decision Process (MDP setup) where the agent has full access to observation $O$, that is, $O=\{x, \hat{x}, \beta, \hat{\beta}\} \in S$. In the second and more difficult setup (POMDP), the environment is partially observable, where $O = \{x, \beta\} \subset  S$. For the latter, the agent cannot observe the state cart velocity and angular velocity information. It has to learn to approximate them using its recurrent connections, and thus needs to retain long-term history~\citep{bakker2002reinforcement,gomez2005co,schafer2008reinforcement}. 
The various RNN-based models are trained to output a distribution over actions at each time step $t$, that is, $\pi(\bm{a}|\bm{s}_{t}, \bm{a}_{t})$, where $\pi$ is the set of policy. For selecting the next action, we consider two policies: (1) sampling: $a_{t+1}\sim \pi(\bm{a}|\bm{s_{t}}, \bm{a_{t}})$, and (2) greedy: $a_{t+1} = \argmax \pi(\bm{a}|\bm{s_{t}}, \bm{a_{t}})$. For all baselines, we use one LSTM layer and one dense layer with softmax to return the distribution over actions. Each layer has 100 hidden units. For VD we tuned the dropout rate $\{0.05, 0.1, 0.2\}$ and for BBB $\mu=\{0, 0.01, 0.05, 0.1\}$ and $\rho=\{-1, -2, -3, -4\}$. For \st we simply set the initial temperature value to $\tau = 1$, the number of possible states to the number of actions ($k=2$), and  the next action is directly selected based on the Gumbel softmax distribution over states. 

Results are presented in Figure \ref{fig:rl}. An important criteria for evaluating RL agents is the sample complexity \citep{malik2019calibrated}, that is, the amount of interactions between agent and environment before a sufficiently good reward is obtained. For both environment setups and both policy types, \st achieves a higher averaged cumulative reward given lower sample complexity. 
Moving from sampling to a greedy policy, \st performance slightly drops. This is easily explained by the inherent sampling process due to the Gumbel softmax. Interestingly, it seems it is this sampling process which allows \st to exhibit a lower sampling complexity in the sampling setups compared to LSTM and VD. In contrast, LSTM and VD show poor performance for the greedy policy setups because they can no longer explore by sampling from the softmax. BBB consistently performs worse than the other methods and we conjecture that this is due to the much larger number of parameters of this model, leading to a worse sampling complexity.
Moving from the MDP to the POMDP, the average accumulative reward naturally drops, as the agents receive less information, but \st again exhibits the best performance for both policy types.

\section{Ablation study on Out-of-Distribution Detection}\label{app:ablation}
\begin{figure}[t!]
	\centering
	\includegraphics[width=0.8\textwidth]{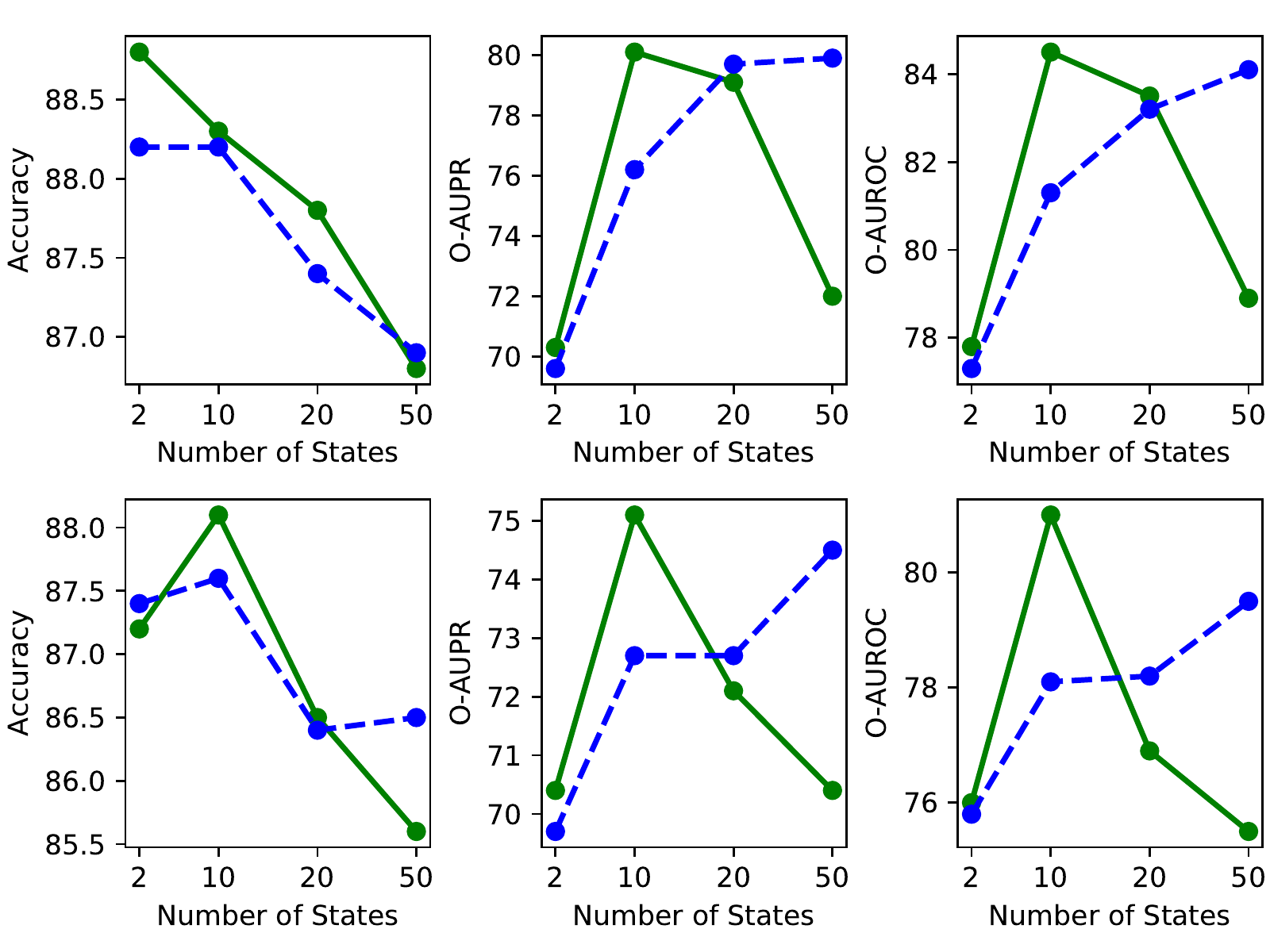}
	\caption{\label{fig:n_states} Apblation study for \st with (1) different number of states (k=2, 10, 20, 50) and (2) learning temperature $\tau$ (green) and fixed temperature $\tau=1$ (blue) on the Out-of-Distribution Detection (OOD) task. The max-probability based O-AUPR and O-AUROC are reported. Top: IMDB(In)/Customer(Out). Bottom: IMDB(In)/Movie(Out). The results are averaged over 10 runs. Learning the temperature $\tau$ is akin to entropy regularization \citep{ szegedy2016rethinking, pereyra2017regularizing,JangETAL:17} and adjusts the confidence (epistemic uncertainty) during training.}
\end{figure}

Figure \ref{fig:n_states} depicts the results of an ablation study focusing on the number of states $k$ and whether or not the temperature $\tau$ was learned. The results are for the two IMDB Out-of-Distribution Detection (OOD) tasks from section~\ref{subsection-ood-experiments} . The results indicate that a smaller number of states is sufficient to capture the \st model's uncertainty on out-of-distribution data. Especially when the temperature parameter is learned during training (the green, solid line), \st shows the best results.
Increasing the number of states of \st, gives the model more capacity to model uncertainty. For 50 states, fixing the temperature to a constant values works better but does not reach the accuracy of \st models with fewer states and learned temperature. 

%% file: paper.bbl
\begin{thebibliography}{60}
\providecommand{\natexlab}[1]{#1}
\providecommand{\url}[1]{\texttt{#1}}
\expandafter\ifx\csname urlstyle\endcsname\relax
  \providecommand{\doi}[1]{doi: #1}\else
  \providecommand{\doi}{doi: \begingroup \urlstyle{rm}\Url}\fi

\bibitem[Abadi et~al.(2015)Abadi, Agarwal, Barham, Brevdo, Chen, Citro,
  Corrado, Davis, Dean, Devin, Ghemawat, Goodfellow, Harp, Irving, Isard, Jia,
  Jozefowicz, Kaiser, Kudlur, Levenberg, Mané, Monga, Moore, Murray, Olah,
  Schuster, Shlens, Steiner, Sutskever, Talwar, Tucker, Vanhoucke, Vasudevan,
  Viégas, Vinyals, Warden, Wattenberg, Wicke, Yu, and Zheng]{45166}
Martín Abadi, Ashish Agarwal, Paul Barham, Eugene Brevdo, Zhifeng Chen, Craig
  Citro, Greg Corrado, Andy Davis, Jeffrey Dean, Matthieu Devin, Sanjay
  Ghemawat, Ian Goodfellow, Andrew Harp, Geoffrey Irving, Michael Isard,
  Yangqing Jia, Rafal Jozefowicz, Lukasz Kaiser, Manjunath Kudlur, Josh
  Levenberg, Dan Mané, Rajat Monga, Sherry Moore, Derek Murray, Chris Olah,
  Mike Schuster, Jonathon Shlens, Benoit Steiner, Ilya Sutskever, Kunal Talwar,
  Paul Tucker, Vincent Vanhoucke, Vijay Vasudevan, Fernanda Viégas, Oriol
  Vinyals, Pete Warden, Martin Wattenberg, Martin Wicke, Yuan Yu, and Xiaoqiang
  Zheng.
\newblock Tensorflow: Large-scale machine learning on heterogeneous distributed
  systems, 2015.

\bibitem[Bakker(2002)]{bakker2002reinforcement}
Bram Bakker.
\newblock {Reinforcement learning with long short-term memory}.
\newblock In \emph{{Advances in Neural Information Processing Systems (NIPS)}},
  2002.

\bibitem[Becker et~al.(2019)Becker, Pandya, Gebhardt, Zhao, Taylor, and
  Neumann]{BeckerPandyaGebhardt2019_1000118248}
Philipp Becker, Harit Pandya, Gregor Gebhardt, Cheng Zhao, C.~James Taylor, and
  Gerhard Neumann.
\newblock Recurrent kalman networks: Factorized inference in high-dimensional
  deep feature spaces.
\newblock In \emph{Proceedings of the 36th International Conference on Machine
  Learning, ICML 2019, 9-15 June 2019, Long Beach, CA, USA}, volume~97 of
  \emph{Proceedings of Machine Learning Research}, pp.\  544–552, 2019.
\newblock URL \url{http://eprints.lincoln.ac.uk/id/eprint/36286/}.

\bibitem[Blundell et~al.(2015)Blundell, Cornebise, Kavukcuoglu, and
  Wierstra]{BlundellETAL:15}
Charles Blundell, Julien Cornebise, Koray Kavukcuoglu, and Daan Wierstra.
\newblock {Weight Uncertainty in Neural Networks}.
\newblock In \emph{{Proceedings of the 32nd International Conference on
  International Conference on Machine Learning (ICML)}}, 2015.

\bibitem[Bridle(1990)]{bridle1990alpha}
John~S Bridle.
\newblock Alpha-nets: a recurrent ‘neural’network architecture with a
  hidden markov model interpretation.
\newblock \emph{Speech Communication}, 9\penalty0 (1):\penalty0 83--92, 1990.

\bibitem[Brockman et~al.(2016)Brockman, Cheung, Pettersson, Schneider,
  Schulman, Tang, and Zaremba]{1606.01540}
Greg Brockman, Vicki Cheung, Ludwig Pettersson, Jonas Schneider, John Schulman,
  Jie Tang, and Wojciech Zaremba.
\newblock {OpenAI Gym}, 2016.

\bibitem[Chung et~al.(2014)Chung, Gulcehre, Cho, and
  Bengio]{chung2014empirical}
Junyoung Chung, Caglar Gulcehre, KyungHyun Cho, and Yoshua Bengio.
\newblock Empirical evaluation of gated recurrent neural networks on sequence
  modeling.
\newblock \emph{arXiv preprint arXiv:1412.3555}, 2014.

\bibitem[Dai et~al.(2017)Dai, Dai, Zhang, Li, and Song]{dai2016recurrent}
Hanjun Dai, Bo~Dai, Yan-Ming Zhang, Shuang Li, and Le~Song.
\newblock Recurrent hidden semi-markov model.
\newblock In \emph{{International Conference on Learning Representations
  (ICLR)}}, 2017.

\bibitem[DeGroot \& Fienberg(1983)DeGroot and Fienberg]{DeGrootFienberg:83}
Morris DeGroot and Stephen Fienberg.
\newblock {The comparison and evaluation of forecasters}.
\newblock \emph{{The Statistician}}, 1983.

\bibitem[Denis \& Esposito(2004)Denis and Esposito]{denis2004learning}
Fran{\c{c}}ois Denis and Yann Esposito.
\newblock Learning classes of probabilistic automata.
\newblock In \emph{International Conference on Computational Learning Theory},
  pp.\  124--139. Springer, 2004.

\bibitem[Depeweg et~al.(2018)Depeweg, Hernandez-Lobato, Doshi-Velez, and
  Udluft]{depeweg2018decomposition}
Stefan Depeweg, Jose-Miguel Hernandez-Lobato, Finale Doshi-Velez, and Steffen
  Udluft.
\newblock Decomposition of uncertainty in bayesian deep learning for efficient
  and risk-sensitive learning.
\newblock In \emph{International Conference on Machine Learning}, pp.\
  1184--1193. PMLR, 2018.

\bibitem[Doerr et~al.(2018)Doerr, Daniel, Schiegg, Duy, Schaal, Toussaint, and
  Sebastian]{doerr2018probabilistic}
Andreas Doerr, Christian Daniel, Martin Schiegg, Nguyen-Tuong Duy, Stefan
  Schaal, Marc Toussaint, and Trimpe Sebastian.
\newblock Probabilistic recurrent state-space models.
\newblock In \emph{International Conference on Machine Learning}, pp.\
  1280--1289, 2018.

\bibitem[Fortunato et~al.(2017)Fortunato, Blundell, and
  Vinyals]{FortunatoETAL:17}
Meire Fortunato, Charles Blundell, and Oriol Vinyals.
\newblock {Bayesian Recurrent Neural Networks}.
\newblock \emph{CoRR}, abs/1704.02798, 2017.

\bibitem[Fraccaro et~al.(2017)Fraccaro, Kamronn, Paquet, and
  Winther]{fraccaro2017disentangled}
Marco Fraccaro, Simon Kamronn, Ulrich Paquet, and Ole Winther.
\newblock A disentangled recognition and nonlinear dynamics model for
  unsupervised learning.
\newblock In \emph{Advances in Neural Information Processing Systems}, pp.\
  3601--3610, 2017.

\bibitem[Gal(2016)]{Gal:16}
Yarin Gal.
\newblock \emph{{Uncertainty in Deep Learning}}.
\newblock PhD thesis, University of Cambridge, 2016.

\bibitem[Gal \& Ghahramani(2016{\natexlab{a}})Gal and
  Ghahramani]{GalGhahramani:16}
Yarin Gal and Zoubin Ghahramani.
\newblock {A Theoretically Grounded Application of Dropout in Recurrent Neural
  Networks}.
\newblock In \emph{{Advances in Neural Information Processing Systems (NIPS)}},
  2016{\natexlab{a}}.

\bibitem[Gal \& Ghahramani(2016{\natexlab{b}})Gal and
  Ghahramani]{GalGhahramani:16b}
Yarin Gal and Zoubin Ghahramani.
\newblock {Dropout as a Bayesian Approximation: Representing Model Uncertainty
  in Deep Learning}.
\newblock In \emph{{Proceedings of The 33rd International Conference on Machine
  Learning (PMLR)}}, New York, New York, USA, 2016{\natexlab{b}}.

\bibitem[Goldberger et~al.(2000)Goldberger, Amaral, Glass, Hausdorff, Ivanov,
  Mark, Mietus, Moody, Peng, and Stanley]{GoldbergerETA:00}
AL~Goldberger, LA~Amaral, L~Glass, JM~Hausdorff, PC~Ivanov, RG~Mark, JE~Mietus,
  GB~Moody, CK~Peng, and HE~Stanley.
\newblock {PhysioBank, PhysioToolkit, and PhysioNet: components of a new
  research resource for complex physiologic signals}.
\newblock \emph{Circulation}, 101\penalty0 (23):\penalty0 E215—20, June 2000.
\newblock ISSN 0009-7322.
\newblock \doi{10.1161/01.cir.101.23.e215}.

\bibitem[Gomez \& Schmidhuber(2005)Gomez and Schmidhuber]{gomez2005co}
Faustino~J Gomez and J{\"u}rgen Schmidhuber.
\newblock Co-evolving recurrent neurons learn deep memory pomdps.
\newblock In \emph{Proceedings of the 7th annual conference on Genetic and
  evolutionary computation}, pp.\  491--498, 2005.

\bibitem[Gumbel(1954)]{Gumbel:54}
Emil~Julius Gumbel.
\newblock {Statistical Theory of Extreme Values and Some Practical
  Applications. A Series of Lectures.}
\newblock \emph{Number 33. US Govt. Print. Office}, 1954.

\bibitem[Guo et~al.(2017)Guo, Pleiss, Sun, and Weinberger]{guo2017calibration}
Chuan Guo, Geoff Pleiss, Yu~Sun, and Kilian~Q Weinberger.
\newblock {On calibration of modern neural networks}.
\newblock In \emph{{Proceedings of the 34th International Conference on Machine
  Learning (ICML)}}, 2017.

\bibitem[Hafner et~al.(2019)Hafner, Lillicrap, Fischer, Villegas, Ha, Lee, and
  Davidson]{hafner2019learning}
Danijar Hafner, Timothy Lillicrap, Ian Fischer, Ruben Villegas, David Ha,
  Honglak Lee, and James Davidson.
\newblock Learning latent dynamics for planning from pixels.
\newblock In \emph{International Conference on Machine Learning}, pp.\
  2555--2565. PMLR, 2019.

\bibitem[Hendrycks \& Gimpel(2017)Hendrycks and Gimpel]{hendrycks2016baseline}
Dan Hendrycks and Kevin Gimpel.
\newblock A baseline for detecting misclassified and out-of-distribution
  examples in neural networks.
\newblock \emph{ICLR}, 2017.

\bibitem[Hinton et~al.(2012)Hinton, Srivastava, Krizhevsky, Sutskever, and
  Salakhutdinov]{HintonETAL:12}
Geoffrey~E. Hinton, Nitish Srivastava, Alex Krizhevsky, Ilya Sutskever, and
  Ruslan Salakhutdinov.
\newblock {Improving neural networks by preventing co-adaptation of feature
  detectors}.
\newblock \emph{CoRR}, abs/1207.0580, 2012.

\bibitem[Hu \& Liu(2004)Hu and Liu]{hu2004mining}
Minqing Hu and Bing Liu.
\newblock Mining and summarizing customer reviews.
\newblock In \emph{Proceedings of the tenth ACM SIGKDD international conference
  on Knowledge discovery and data mining}, pp.\  168--177, 2004.

\bibitem[H{\"u}llermeier \& Waegeman(2019)H{\"u}llermeier and
  Waegeman]{hullermeier2019aleatoric}
Eyke H{\"u}llermeier and Willem Waegeman.
\newblock Aleatoric and epistemic uncertainty in machine learning: A tutorial
  introduction.
\newblock \emph{arXiv preprint arXiv:1910.09457}, 2019.

\bibitem[Hwang et~al.(2020)Hwang, Mehta, Kim, Johnson, and
  Singh]{hwang2020sampling}
Seong~Jae Hwang, Ronak~R Mehta, Hyunwoo~J Kim, Sterling~C Johnson, and Vikas
  Singh.
\newblock Sampling-free uncertainty estimation in gated recurrent units with
  applications to normative modeling in neuroimaging.
\newblock In \emph{Uncertainty in Artificial Intelligence}, pp.\  809--819.
  PMLR, 2020.

\bibitem[Jang et~al.(2017)Jang, Gu, and Poole]{JangETAL:17}
Eric Jang, Shixiang Gu, and Ben Poole.
\newblock {Categorical Reparameterization with Gumbel-Softmax}.
\newblock In \emph{{5th International Conference on Learning Representations
  (ICLR)}}, 2017.

\bibitem[Jiang et~al.(2018)Jiang, Kim, Guan, and Gupta]{jiang2018trust}
Heinrich Jiang, Been Kim, Melody Guan, and Maya Gupta.
\newblock To trust or not to trust a classifier.
\newblock In \emph{Advances in neural information processing systems}, pp.\
  5541--5552, 2018.

\bibitem[{Kachuee} et~al.(2018){Kachuee}, {Fazeli}, and
  {Sarrafzadeh}]{KachueeETAL:18}
M.~{Kachuee}, S.~{Fazeli}, and M.~{Sarrafzadeh}.
\newblock {ECG Heartbeat Classification: A Deep Transferable Representation}.
\newblock In \emph{{IEEE International Conference on Healthcare Informatics
  (ICHI)}}, 2018.

\bibitem[Kendall \& Gal(2017)Kendall and Gal]{KendallGal:17}
Alex Kendall and Yarin Gal.
\newblock {What Uncertainties Do We Need in Bayesian Deep Learning for Computer
  Vision?}
\newblock In \emph{{Advances in Neural Information Processing Systems (NIPS)}},
  2017.

\bibitem[Kingma \& Ba(2015)Kingma and Ba]{adam}
Diederick~P Kingma and Jimmy Ba.
\newblock {Adam: A method for stochastic optimization}.
\newblock In \emph{{International Conference on Learning Representations
  (ICLR)}}, 2015.

\bibitem[Krakovna \& Doshi-Velez(2016)Krakovna and
  Doshi-Velez]{krakovna2016increasing}
Viktoriya Krakovna and Finale Doshi-Velez.
\newblock Increasing the interpretability of recurrent neural networks using
  hidden markov models.
\newblock \emph{arXiv preprint arXiv:1606.05320}, 2016.

\bibitem[Krzywinski \& Altman(2013)Krzywinski and Altman]{KrzywinskiAltman:13}
Martin Krzywinski and Naomi Altman.
\newblock {Points of significance: Importance of being uncertain}.
\newblock \emph{Nature Methods}, 10\penalty0 (9):\penalty0 809--810, 2013.
\newblock ISSN 1548-7091.
\newblock \doi{10.1038/nmeth.2613}.

\bibitem[Kuleshov \& Liang(2015)Kuleshov and Liang]{kuleshov2015calibrated}
Volodymyr Kuleshov and Percy~S Liang.
\newblock Calibrated structured prediction.
\newblock In \emph{Advances in Neural Information Processing Systems}, pp.\
  3474--3482, 2015.

\bibitem[Kuleshov et~al.(2018)Kuleshov, Fenner, and Ermon]{KuleshovETAL:18}
Volodymyr Kuleshov, Nathan Fenner, and Stefano Ermon.
\newblock {Accurate Uncertainties for Deep Learning Using Calibrated
  Regression}.
\newblock In \emph{{Proceedings of the 35th International Conference on Machine
  Learning (ICML)}}, 2018.

\bibitem[Kwon et~al.(2018)Kwon, Won, Kim, and Paik]{KwonETAL:18}
Yongchan Kwon, Joong-Ho Won, Beom~Joon Kim, and Myunghee~Cho Paik.
\newblock {Uncertainty quantification using Bayesian neural networks in
  classification: Application to ischemic stroke lesion segmentation}.
\newblock In \emph{{International Conference on Medical Imaging with Deep
  Learning (MIDL)}}, 2018.

\bibitem[Lakshminarayanan et~al.(2017)Lakshminarayanan, Pritzel, and
  Blundell]{LakshminarayananETAL:17}
Balaji Lakshminarayanan, Alexander Pritzel, and Charles Blundell.
\newblock {Simple and Scalable Predictive Uncertainty Estimation using Deep
  Ensembles}.
\newblock In \emph{{Advances in Neural Information Processing Systems (NIPS)}},
  2017.

\bibitem[Linderman et~al.(2017)Linderman, Johnson, Miller, Adams, Blei, and
  Paninski]{linderman2017bayesian}
Scott Linderman, Matthew Johnson, Andrew Miller, Ryan Adams, David Blei, and
  Liam Paninski.
\newblock Bayesian learning and inference in recurrent switching linear
  dynamical systems.
\newblock In \emph{Artificial Intelligence and Statistics}, pp.\  914--922,
  2017.

\bibitem[Maas et~al.(2011)Maas, Daly, Pham, Huang, Ng, and Potts]{MaasETAL:11}
Andrew~L. Maas, Raymond~E. Daly, Peter~T. Pham, Dan Huang, Andrew~Y. Ng, and
  Christopher Potts.
\newblock {Learning Word Vectors for Sentiment Analysis}.
\newblock In \emph{{Proceedings of the 49th Annual Meeting of the Association
  for Computational Linguistics: Human Language Technologies (NAACL)}}, 2011.

\bibitem[Maddison et~al.(2017)Maddison, Mnih, and Teh]{MaddisonETAL:17}
Chris~J. Maddison, Andriy Mnih, and Yee~Whye Teh.
\newblock {The Concrete Distribution: {A} Continuous Relaxation of Discrete
  Random Variables}.
\newblock In \emph{{5th International Conference on Learning Representations
  (ICLR)}}, 2017.

\bibitem[Malik et~al.(2019)Malik, Kuleshov, Song, Nemer, Seymour, and
  Ermon]{malik2019calibrated}
Ali Malik, Volodymyr Kuleshov, Jiaming Song, Danny Nemer, Harlan Seymour, and
  Stefano Ermon.
\newblock Calibrated model-based deep reinforcement learning.
\newblock In \emph{International Conference on Machine Learning}, pp.\
  4314--4323, 2019.

\bibitem[{Moody} \& {Mark}(2001){Moody} and {Mark}]{MoodyMark:01}
G.~B. {Moody} and R.~G. {Mark}.
\newblock {The impact of the MIT-BIH Arrhythmia Database}.
\newblock \emph{IEEE Engineering in Medicine and Biology Magazine}, 20\penalty0
  (3):\penalty0 45--50, May 2001.
\newblock ISSN 0739-5175.
\newblock \doi{10.1109/51.932724}.

\bibitem[Naeini et~al.(2015)Naeini, Cooper, and Hauskrecht]{NaeiniETAL:15}
Mahdi~Pakdaman Naeini, Gregory~F. Cooper, and Milos Hauskrecht.
\newblock {Obtaining well calibrated probabilities using bayesian binning}.
\newblock In \emph{{Proceedings of the Twenty-Ninth AAAI Conference on
  Artificial Intelligence (AAAI)}}, 2015.

\bibitem[Niculescu-Mizil \& Caruana(2005)Niculescu-Mizil and
  Caruana]{Niculescu-MizilETAL:05}
Alexandru Niculescu-Mizil and Rich Caruana.
\newblock {Predicting Good Probabilities with Supervised Learning}.
\newblock In \emph{{Proceedings of the 22nd International Conference on Machine
  Learning (ICML)}}, 2005.

\bibitem[Noreen(1989)]{Noreen:1989}
Eric~W. Noreen.
\newblock \emph{{Computer Intensive Methods for Testing Hypotheses: An
  Introduction}}.
\newblock Wiley, New York, 1989.

\bibitem[Pang et~al.(2002)Pang, Lee, and Vaithyanathan]{pang-etal-2002-thumbs}
Bo~Pang, Lillian Lee, and Shivakumar Vaithyanathan.
\newblock Thumbs up? sentiment classification using machine learning
  techniques.
\newblock In \emph{Proceedings of the 2002 Conference on Empirical Methods in
  Natural Language Processing ({EMNLP} 2002)}, pp.\  79--86. Association for
  Computational Linguistics, July 2002.

\bibitem[Pereyra et~al.(2017)Pereyra, Tucker, Chorowski, Kaiser, and
  Hinton]{pereyra2017regularizing}
Gabriel Pereyra, George Tucker, Jan Chorowski, {\L}ukasz Kaiser, and Geoffrey
  Hinton.
\newblock {Regularizing neural networks by penalizing confident output
  distributions}.
\newblock \emph{arXiv preprint arXiv:1701.06548}, 2017.

\bibitem[Platt et~al.(1999)]{platt1999probabilistic}
John Platt et~al.
\newblock Probabilistic outputs for support vector machines and comparisons to
  regularized likelihood methods.
\newblock \emph{Advances in large margin classifiers}, 10\penalty0
  (3):\penalty0 61--74, 1999.

\bibitem[Rabin(1963)]{rabin1963probabilistic}
Michael~O Rabin.
\newblock {Probabilistic automata}.
\newblock \emph{{Information and Control}}, 6\penalty0 (3):\penalty0 230--245,
  1963.

\bibitem[Sch{\"a}fer(2008)]{schafer2008reinforcement}
Anton~Maximilian Sch{\"a}fer.
\newblock {Reinforcement learning with recurrent neural networks}, 2008.

\bibitem[Schellhammer et~al.(1998)Schellhammer, Diederich, Towsey, and
  Brugman]{Schellhammer:1998}
Ingo Schellhammer, Joachim Diederich, Michael Towsey, and Claudia Brugman.
\newblock {Knowledge Extraction and Recurrent Neural Networks: An Analysis of
  an Elman Network Trained on a Natural Language Learning Task}.
\newblock In \emph{{Proceedings of the Joint Conferences on New Methods in
  Language Processing and Computational Natural Language Learning}}, 1998.

\bibitem[Srivastava et~al.(2014)Srivastava, Hinton, Krizhevsky, Sutskever, and
  Salakhutdinov]{SrivastavaETAL:14}
Nitish Srivastava, Geoffrey Hinton, Alex Krizhevsky, Ilya Sutskever, and Ruslan
  Salakhutdinov.
\newblock {Dropout: A Simple Way to Prevent Neural Networks from Overfitting}.
\newblock \emph{Journal of Machine Learning Research}, 15:\penalty0 1929--1958,
  2014.

\bibitem[Szegedy et~al.(2016)Szegedy, Vanhoucke, Ioffe, Shlens, and
  Wojna]{szegedy2016rethinking}
Christian Szegedy, Vincent Vanhoucke, Sergey Ioffe, Jon Shlens, and Zbigniew
  Wojna.
\newblock {Rethinking the inception architecture for computer vision}.
\newblock In \emph{{Proceedings of the IEEE conference on computer vision and
  pattern recognition}}, 2016.

\bibitem[Tomita(1982)]{tomita:cogsci82}
M.~Tomita.
\newblock Dynamic construction of finite automata from examples using
  hill-climbing.
\newblock In \emph{{P}roceedings of the Fourth Annual Conference of the
  Cognitive Science Society}, pp.\  105--108, 1982.

\bibitem[Wang \& Niepert(2019)Wang and Niepert]{WangNiepert:19}
Cheng Wang and Mathias Niepert.
\newblock {State-Regularized Recurrent Neural Networks}.
\newblock In \emph{{Proceedings of the 36th International Conference on Machine
  Learning (ICML)}}, 2019.

\bibitem[Wang et~al.(2018)Wang, Zhang, Ororbia~II, Xing, Liu, and
  Giles]{Wang:2007:nc}
Qinglong Wang, Kaixuan Zhang, Alexander~G. Ororbia~II, Xinyu Xing, Xue Liu, and
  C.~Lee Giles.
\newblock {An Empirical Evaluation of Rule Extraction from Recurrent Neural
  Networks}.
\newblock \emph{Neural Computation}, 30\penalty0 (9):\penalty0 2568--2591,
  2018.

\bibitem[Weiss et~al.(2018)Weiss, Goldberg, and Yahav]{pmlr-v80-weiss18a}
Gail Weiss, Yoav Goldberg, and Eran Yahav.
\newblock {Extracting Automata from Recurrent Neural Networks Using Queries and
  Counterexamples}.
\newblock In \emph{{Proceedings of the 35th International Conference on Machine
  Learning}}, Proceedings of Machine Learning Research, 2018.

\bibitem[Weiss et~al.(2019)Weiss, Goldberg, and Yahav]{WeissETAL:19}
Gail Weiss, Yoav Goldberg, and Eran Yahav.
\newblock {Learning Deterministic Weighted Automata with Queries and
  Counterexamples}.
\newblock In \emph{{Advances in Neural Information Processing Systems 32}},
  pp.\  8558--8569. Curran Associates, Inc., 2019.

\bibitem[Zadrozny \& Elkan(2002)Zadrozny and Elkan]{zadrozny2002transforming}
Bianca Zadrozny and Charles Elkan.
\newblock Transforming classifier scores into accurate multiclass probability
  estimates.
\newblock In \emph{Proceedings of the eighth ACM SIGKDD international
  conference on Knowledge discovery and data mining}, pp.\  694--699, 2002.

\end{thebibliography}
